\documentclass{jair}
\usepackage{hyperref}

\usepackage{amsmath}
\usepackage{amsthm}
\usepackage{algorithm}
\usepackage{algpseudocode}
\usepackage{adjustbox}
\usepackage{longtable}
\usepackage{booktabs}
\usepackage{placeins}
\usepackage{natbib}

\usepackage{microtype}
\newtheorem{definition}{Definition}

\usepackage{amssymb}
\usepackage{graphicx} 
\usepackage{xcolor}
\usepackage{booktabs} 
\usepackage{multirow}

\usepackage{threeparttable}
\usepackage{url}

\usepackage{xcolor}

\newcommand{\yn}[1]{\textcolor{black}{#1}}

\newcommand{\hz}[1]{\textcolor{black}{#1}}
\newcommand{\hzz}[1]{\textcolor{black}{#1}}
\newcommand{\hzzz}[1]{\textcolor{black}{#1}}

\newcommand{\RA}[1]{\textcolor{black}{#1}}
\newcommand{\RB}[1]{\textcolor{black}{#1}}
\newcommand{\RC}[1]{\textcolor{black}{#1}}
\newcommand{\RALL}[1]{\textcolor{black}{#1}}
\newcommand{\RABC}[1]{\textcolor{black}{#1}}

\newcommand{\FN}[1]{\textcolor{black}{#1}}

\setcopyright{cc}
\copyrightyear{2026}
\acmDOI{10.1613/jair.1.20611}

\JAIRAE{Sven Koenig}
\JAIRTrack{Multi-Agent Path Finding}
\acmVolume{85}
\acmArticle{28}
\acmMonth{3}
\acmYear{2026}

\begin{document}

\title{Learning-guided Prioritized Planning for Lifelong Multi-Agent Path Finding in Warehouse Automation}

\author{Han Zheng}
\email{hanzheng@mit.edu}
\orcid{0009-0008-3768-5556}
\affiliation{%
  \institution{Massachusetts Institute of Technology}
  \city{Cambridge}
  \state{MA}
  \country{USA}
}

\author{Yining Ma}
\authornote{Corresponding Author.}
\orcid{0000-0002-6639-8547}
\email{yiningma@mit.edu}
\affiliation{%
  \institution{Massachusetts Institute of Technology}
  \city{Cambridge}
  \state{MA}
  \country{USA}
}
\author{Brandon Araki}
\orcid{0000-0002-3094-1587}
\email{maraki@symbotic.com}
\affiliation{%
  \institution{Symbotic}
  \city{Wilmington}
  \state{MA}
  \country{USA}
}

\author{Jingkai Chen}
\orcid{0000-0002-3528-8185}
\email{jichen@symbotic.com}
\affiliation{%
  \institution{Symbotic}
  \city{Wilmington}
  \state{MA}
  \country{USA}
}
\author{Cathy Wu}
\orcid{0000-0001-8594-303X}
\email{cathywu@mit.edu}
\affiliation{%
  \institution{Massachusetts Institute of Technology}
  \city{Cambridge}
  \state{MA}
  \country{USA}
}

\renewcommand{\shortauthors}{Zheng, Ma, Araki, Chen \& Wu}

\begin{abstract}
Lifelong Multi-Agent Path Finding (MAPF) is critical for modern warehouse automation, which requires multiple robots to continuously navigate conflict-free paths to optimize the overall system throughput. However, the complexity of warehouse environments \RC{and the long-term dynamics of lifelong MAPF} often demand costly adaptations to classical search-based solvers. While machine learning methods have been explored, their superiority over search-based methods remains inconclusive. In this paper, we introduce Reinforcement Learning (RL) guided Rolling Horizon Prioritized Planning (RL-RH-PP), the first framework integrating RL with search-based planning for lifelong MAPF. Specifically, we leverage classical Prioritized Planning (PP) as a backbone for its simplicity and flexibility in integrating with a learning-based priority assignment policy. By formulating dynamic priority assignment as a Partially Observable Markov Decision Process (POMDP), RL-RH-PP exploits the sequential decision-making nature of lifelong planning while delegating complex spatial-temporal interactions among agents to reinforcement learning. An attention-based neural network autoregressively decodes priority orders on-the-fly, enabling efficient sequential single-agent planning by the PP planner. Evaluations in realistic warehouse simulations show that RL-RH-PP achieves the highest total throughput among baselines and \RALL{generalizes effectively across agent densities, planning horizons, and warehouse layouts}. Our interpretive analyses reveal that RL-RH-PP proactively prioritizes congested agents and strategically redirects agents from congestion, easing traffic flow and boosting throughput. These findings highlight the potential of learning-guided approaches to augment traditional heuristics in modern warehouse automation.
\end{abstract}

\received{30 September 2025}
\received[accepted]{04 February 2026}
\maketitle

\section{Introduction}
\label{sec:intro}
\begin{figure*}[t!]
    \centering
    \includegraphics[width=0.5\textwidth]{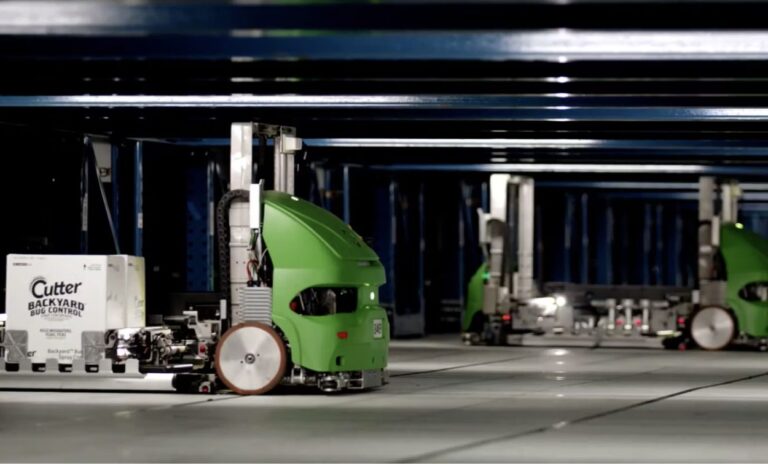}
    \caption{Robot fleet routing in a Symbotic warehouse (source: \href{https://www.symbotic.com}{https://www.symbotic.com}).}
\label{fig:symbotic real}
\end{figure*}

The rapid growth of e-commerce has intensified the demand for efficient, high-throughput logistics systems \citep{Yu2016commerce}. Traditional manual workflows struggle with scalability and high labor costs \citep{ivinjak2022CasestudyAO}. To address these challenges, warehouse automation has transformed modern logistics by leveraging recent advances in robotics, optimization, and artificial intelligence (AI) to streamline operations and enhance efficiency \citep{Warehousereview}. Automated warehouse systems such as sortation and fulfillment centers have evolved into large-scale fleets of autonomous mobile robots (AMRs) navigating complex warehouse layouts \citep{Wurman2007CoordinatingHO}. As shown in Figure~\ref{fig:symbotic real}, Symbotic, a leading warehouse automation company, orchestrates advanced AMRs in high-density systems, reinventing warehouse automation for increased efficiency. In such AMR systems, effective and sustained coordination is essential to alleviate congestion, reduce operational costs, and maximize throughput. Even small improvements in AMR coordination can lead to substantial efficiency gains, boosting throughput while driving down costs \citep{Farinelli2016,Atzmon2020}.

\hz{A central paradigm for coordinating AMRs in warehouses is the Multi-Agent Path Finding (MAPF) problem \citep{stern2019multi},} where multiple agents navigate from start positions to assigned goals while avoiding conflicts. MAPF aims to optimize global objectives such as minimizing total travel time (flowtime) or the completion time of the last agent (makespan). Although originally developed for warehouse automation, it also has broad applications in traffic coordination \citep{yan2024multi,Zheng2024}, airport logistics \citep{MAPFAriport}, and video games \citep{DeWilde2014,Sigurdson2018}. 

\RC{Classic MAPF assumes a static scenario with predefined goals and performs one-shot planning.} Various methods have been proposed to solve one-shot MAPF, ranging from optimal methods like \hz{Conflict-Based Search (CBS) \citep{sharon2015conflict}} to heuristics that trade optimality for speed \citep{ECBS,ma2019searching,PIBT}. Among these, Prioritized Planning (PP) \citep{erdmann1987multiple} is a simple yet powerful heuristic that decomposes MAPF into sequential single-agent path plans based on a predefined priority order. With a well-chosen order, PP offers impressive scalability and efficiency in large, dynamic environments.

\RC{Extending beyond one-shot MAPF, real-world scenarios require a more dynamic formulation over the time dimension, known as lifelong MAPF \citep{ma2017lifelong}, where agents are continuously assigned new tasks after completing their current ones. This formulation is crucial for modern warehouse automation, such as fulfillment centers and sortation hubs, where robots must transition seamlessly between tasks to maximize overall throughput continuously \citep{liu2019task}. Unlike one-shot MAPF, lifelong MAPF introduces several new challenges  \citep{PIBT}: (i) agents repeatedly enter and leave the system, leading to continual re-coordination across time, (ii) congestion patterns evolve dynamically as tasks accumulate, requiring foresight rather than static planning, and (iii) myopic decisions can create cascading inefficiencies, as early plans directly shape future feasibility; (iv) paths with good quality need to be found within a realistic timeframe.}  \RC{Those challenges distinguish lifelong MAPF from one-shot MAPF and motivate the need for frameworks that explicitly model long-horizon interactions while adapting to shifting conditions.}

Recently, machine learning (ML) techniques have shown promise in helping address real-world complexities for one-shot MAPF \citep{Primal,huang2021learning,yan2024neural}. However, they have yet to consistently outperform search-based methods in the more complex yet less explored lifelong settings \citep{Primal2,Follower}. \RC{As mentioned above, the key challenge is that planning in lifelong MAPF is inherently a sequential decision-making process with causal dependencies, where each planning decision directly influences future planning outcomes. Thus, approaches that only optimize for immediate conflict avoidance or short-term efficiency-as is often sufficient in one-shot MAPF-are insufficient in the lifelong setting, since they may worsen long-term congestion or deadlock risk.} Effective solutions must account for these cascading effects by maintaining a longer preview horizon to explicitly model how current decisions shape future task allocations and path interactions. \RB{This requirement complicates the direct adaptation of straightforward one-shot or short-horizon learning-based approaches, which often struggle with efficiently handling long-term dependencies and dynamic task assignment.} This presents an exciting opportunity to leverage deep reinforcement learning, in novel ways that explicitly capture the causal structure of lifelong MAPF, potentially enabling learning-based methods to surpass search-based techniques in such dynamic, real-world settings.

In this paper, we introduce \textbf{RL-guided Rolling Horizon Prioritized Planning (RL-RH-PP)}, the first hybrid framework that integrates reinforcement learning (RL) for dynamic priority order generation with search-based prioritized planning (PP). \RC{Our design choices are tailored to lifelong MAPF: the rolling horizon mechanism enables continual re-planning as new tasks arrive, while the RL-guided priority assignment explicitly captures long-horizon dependencies and adapts prioritization to evolving congestion patterns. Our motivation is to leverage reinforcement learning to address the hard-to-model sequential decision-making challenges of lifelong MAPF by capturing complex spatiotemporal interactions among agents, while integrating a simple yet efficient prioritized planning (PP) framework to achieve the best of both worlds. Importantly, PP provides an attractive backbone because it is lightweight, highly scalable, and easily extensible to dynamic settings, unlike more complex search-based solvers (e.g., CBS or PBS) whose computational cost grows rapidly with team size. The sequential structure of rolling horizon planning is naturally compatible with learning-guided optimization, as it exposes the long-horizon ordering decision as the key lever of coordination. This synergy allows us to focus the learning component on optimizing priorities, while relying on PP’s efficiency to compute collision-free paths in real time, making the overall framework achieve the best of both worlds.}

Specifically, RL-RH-PP casts dynamic priority assignment as a Partially Observable Markov Decision Process (POMDP) and uses a learned neural network to autoregressively decode priority orders. These orders are then fed into a rolling-horizon extension of Prioritized Planning (RH-PP), where multiple promising priority orders are sampled, with a conflict-repair mechanism applied as needed to ensure high-quality, conflict-free path planning. By integrating data-driven RL with PP, our learning-guided RL-RH-PP induces a dynamically streamlined search space for priority order sampling, significantly boosting throughput for lifelong MAPF, especially in complex, dynamic warehouse environments.

\hz{At the core of our proposed RL-RH-PP is a transformer-styled neural network \citep{vaswani2017attention} that captures both \textit{temporal} and \textit{spatial} dependencies in multi-agent interactions for lifelong MAPF.} Specifically, the proposed encoder processes multiple agent paths spanning the time dimension using a dictionary-based embedding mechanism and stacked multi-head attention layers with alternating temporal and spatial attention mechanisms. The temporal attention captures long-term dependencies along each agent's path over time, while the spatial attention models dynamic interactions between agents in the warehouse map. This enables RL-RH-PP to dynamically adjust prioritization in response to changing navigation constraints. Based on the learned agent embeddings, the decoder then autoregressively constructs the total priority orders for RH-PP execution, where diverse promising priority orders can be efficiently sampled from such a learned policy. By integrating spatial and temporal context in a data-driven manner, RL-RH-PP generates high-quality priority orders that account for both current and future agent interactions,  which significantly improves the agent coordination and throughput, especially in dense environments with a high likelihood of congestion.

We train RL-RH-PP in a warehouse simulation environment built on real-world-inspired warehouse such as Amazon fulfillment center and Symbotic warehouses. \hzz{Our work is the first to incorporate the Symbotic warehouse layout into lifelong MAPF research, whereas prior studies have predominantly focused on Amazon warehouse maps \citep{RHCR,jiang2024scalinglifelongmultiagentpath}. The Symbotic warehouse presents a fundamentally different structure, characterized by higher obstacle density, structured regional constraints, and distinct navigation challenges such as bottlenecks and congestion-prone cross-aisles. This contribution provides new benchmarking opportunities and helps validate lifelong MAPF methods in more constrained and operationally relevant settings.}

The RL policy interacts with an RH-PP planner, observes state transitions, and receives feedback through a customized reward function. The reward incentivizes efficient lifelong MAPF by minimizing robust travel distances and reducing congestion.
Extensive experiments demonstrate that RL-RH-PP significantly improves throughput and planning efficiency, achieving on average 25\% higher throughput over RH-PP with random priority orders. \RB{Furthermore, RL-RH-PP exhibits strong zero-shot generalization across varying agent densities, planning horizons, and unseen maps, achieving the highest total throughput over various search-based baselines, especially in scenarios with high obstacle density.} \hzz{Experiments and analyses, including rendering priority heatmaps and agent movement traces, show that RL-RH-PP learns to assign higher priorities to agents in congested regions, alleviating bottlenecks and quickly recovering from severe deadlocks. We illustrate how RL-RH-PP steers agents away from congested regions, opening pathways for other robots and improving long-term throughput, making RL-RH-PP a robust solution for more efficient path planning in complex, dynamic warehouse automation environments.}

To summarize, our key contributions are:
\begin{itemize}
    \item We propose RL-RH-PP, the first deep reinforcement learning-guided framework that dynamically optimizes priority orders for lifelong MAPF.
    \item We introduce RH-PP, a rolling-horizon extension of Prioritized Planning (PP) that serves as the efficient backbone for learning-guided decision-making.
    \item We design a transformer-style neural architecture that captures both spatial and temporal dependencies, allowing for data-driven priority order optimization.
    \item We achieve gains in throughput, planning efficiency, and generalization, demonstrating RL-RH-PP's superiority over various baselines in solving the challenging lifelong MAPF.
    \item \hzz{We conduct interpretable analysis using priority heatmaps and agent-trace comparisons, revealing how RL-RH-PP prioritizes congested agents and recovers from deadlocks and boosts throughput.}
\end{itemize}

Notably, our findings highlight the power of learning-guided approaches in advancing multi-agent coordination in warehouse automation and beyond. The RL-RH-PP framework and training pipeline will be open-sourced to facilitate further research at \url{https://github.com/MikeZheng777/RL-RH-PP}.

\RALL{The remainder of this paper is structured as follows. Section~\ref{sec:related work} discusses related work on MAPF and lifelong MAPF, highlighting existing learning-based approaches. Section~\ref{sec:problem_def} formally defines the MAPF and Lifelong MAPF problems. Section~\ref{sec:Method} introduces our rolling-horizon extension of PP, and formulates the lifelong priority assignment as a POMDP and presents the solution RL-RH-PP. Section~\ref{sec:experiments} presents the evaluation and generalization behavior of RL-RH-PP and reports ablation studies on architectural and algorithmic choices. Section~\ref{sec:interpretatoin} provides an interpretation of the learned policy. We conclude with a summary of findings and future research directions in Section~\ref{sec:conclusion}.}

\hzz{}

\section{Related Work}
\label{sec:related work}
We now discuss the bodies of work that form the foundation of our method and highlight how our approach differs from existing techniques.

\subsection{Multi-Agent Path Finding}
Classical multi-agent path finding (MAPF) \citep{stern2019multi} is NP-hard \citep{yu2013structure} and involves computing conflict-free paths for multiple agents on a discrete graph. Due to the vast joint trajectory space \citep{sharon2015conflict}, most MAPF algorithms rely on single-agent planners such as A* \citep{hart1968formal} and SIPP \citep{phillips2011sipp}, treating other agents' paths as constraints. Conflict-Based Search (CBS) \citep{sharon2015conflict} remains the state-of-the-art optimal MAPF algorithm, resolving conflicts via backtracking while ensuring completeness and optimality; however, its exponential worst-case complexity limits scalability. Enhanced CBS (ECBS) \citep{barer2014suboptimal} improves efficiency by introducing bounded suboptimality, balancing speed and solution quality. For greater scalability, Prioritized Planning (PP) \citep{erdmann1987multiple} and Priority-Based Search (PBS) \citep{ma2019searching} plan sequentially based on agent priorities, treating higher-priority agents as constraints. While PP is efficient, it lacks completeness and optimality guarantees; PBS mitigates this by searching over global priority orders. Priority Inheritance with Backtracking (PIBT) \citep{PIBT} further improves adaptability by dynamically adjusting priorities to prevent deadlocks, though it sacrifices optimality due to localized decision-making. 

Beyond search-based families, one-shot MAPF has also advanced through powerful repair and search techniques. \FN{MAPF-LNS2 \citep{LiAAAI22} performs large-neighborhood search with fast repairing, yielding strong anytime behavior. LaCAM* \citep{okumura2024engineeringlacamastrealtimelargescale} targets real-time, large-scale settings with near-optimal performance via careful engineering of cardinal conflict handling. Compared to these approaches, prioritized planning remains a compelling choice for settings demanding fast and scalable solutions. Our contribution is orthogonal: we keep a lightweight prioritized-planning backbone for its speed and efficiency and learn to select effective global priority orders within to improve solution quality under lifelong planning.}

\subsection{Lifelong Multi-Agent Path Finding}
Lifelong MAPF algorithms balance solution quality and computational efficiency, with different approaches excelling in distinct scenarios. Rolling-Horizon Conflict Resolution (RHCR) \citep{RHCR} introduces a planning window to manage complexity. It integrates well with centralized planners like CBS \citep{sharon2015conflict} and PBS \citep{ma2019searching}, enabling high-quality solutions in small but dense environments; however, RHCR struggles with scalability when handling thousands of agents due to its computational overhead. In contrast, PIBT \citep{PIBT} offers a highly scalable alternative, employing a greedy single-step decentralized approach that achieves real-time performance even with thousands of agents, albeit with solution-quality trade-offs due to its local decisions. \FN{Our method follows a centralized rolling-horizon formulation where we use RL to propose a high-quality global priority order, rather than making purely local one-step decisions as in PIBT.}

To better balance speed and quality, WPPL \citep{jiang2024scalinglifelongmultiagentpath}---winner of the 2023 League of Robot Runner Competition \citep{chan2024the}---combines PIBT for fast initial plans with windowed MAPF-LNS \citep{li2021anytime} for refinement. Our approach shares the rolling-horizon concept with RHCR and prioritization strategies with PP \citep{erdmann1987multiple} but differs by leveraging reinforcement learning to dynamically optimize global priority orders within each window.

\subsection{Learning-based Approaches for MAPF and Lifelong MAPF}
A growing body of work integrates learning with search to improve scalability and solution quality. For one-shot MAPF, \cite{huang2021learning} employ a linear model with handcrafted features to guide CBS, showing that even simple predictors can assist traditional solvers. PRIMAL \citep{Primal} uses imitation and multi-agent RL for decentralized coordination based on local neighborhoods encoded by CNNs. Transformer-based models \citep{yan2024neural} improve MAPF-LNS \citep{li2021anytime} by learning powerful heuristics that scale to thousands of agents, \RC{while ALPHA \citep{he2023alphaattentionbasedlonghorizonpathfinding} targets attention-based long-horizon planning in structured environments. }

\RC{Several recent studies further explore how learning and search can be synergistic. \cite{veerapaneni2024improving} shows that augmenting learnt local policies with heuristic search yields consistent gains, and \cite{WangAAAI25} combines multi-agent RL with LNS, reinforcing the utility of hybrid pipelines that couple global repair with learned guidance. On the learning-rule side, simple imitation of CS-PIBT can outperform large-scale imitation learning for MAPF \citep{veerapaneni2024worksmarterhardersimple}, and social-behavior priors have been shown to improve coordination under dilemma-like interactions \citep{he2024socialbehaviorkeylearningbased}.}

\RC{For lifelong MAPF specifically, PRIMAL$_2$ \citep{Primal2} and Followers \citep{Follower} use RL to provide global guidance over extended horizons. More recently, large-scale imitation learning has been shown to scale lifelong MAPF to ten thousand robots \citep{jiang2025deployingthousandrobotsscalable}, highlighting the promise of data-driven coordination at warehouse scale. Despite these advances, learning-based approaches have not universally displaced strong non-learning baselines such as RHCR \citep{RHCR} and LNS variants \citep{LiAAAI22,li2021anytime, jiang2024scalinglifelongmultiagentpath}. These observations motivate our evaluation protocol and the design of RL-RH-PP: rather than replacing search, we learn to propose high-quality global priority orders that a rolling-horizon PP backbone executes efficiently.}

\RABC{
\subsection{Prioritized Planning}
Prioritized Planning (PP) was first proposed in \cite{erdmann1987multiple}. It is a classic decoupled MAPF approach that assigns a strict priority ordering to agents and plans their paths sequentially, where each agent avoids conflicts with all higher-priority agents. While PP does not guarantee completeness or optimality, it remains popular for its efficiency and simplicity, with performance hinging on how the priority ordering is determined. Heuristic strategies include the query-distance heuristic \citep{Berg2005,ma2019searching}, which prioritizes agents by travel distance, the least-option heuristic \citep{wu2019}, which favors agents with fewer feasible path options. When no strong heuristic is available, random priority orderings combined with random restarts can improve success rates \citep{Bennewitz2002}. }

\RABC{Most closely related to our approach is \cite{ZhangSoCS22}, who introduced a machine learning framework to automatically learn priority orderings for PP for one-shot MAPF. Their approach extracts handcrafted features, and trains linear ranking models to predict agent priorities by supervised learning. While the ML-guided PP proposed in \cite{ZhangSoCS22} achieved moderate improvements over random priority ordering, it struggled to consistently outperform strong heuristics such as the query-distance heuristic \citep{Berg2005}, reflecting the limitations of feature engineering and static supervision. Our work differs fundamentally: 1) We study lifelong MAPF with continual task arrivals, whereas \cite{ZhangSoCS22} addresses the one-shot MAPF setting. 2) rather than relying on predefined features and supervised learning, we formulate dynamic priority assignment as a reinforcement learning problem. By integrating an attention-based neural policy, our approach adapts priorities online to congestion and spatial-temporal interactions, yielding significantly higher throughput in lifelong MAPF.}

\section{Problem Definition}
\label{sec:problem_def}
\begin{definition}[MAPF] 
The \textit{Multi-Agent Path Finding (MAPF)} problem involves computing conflict-free paths for a set of \( N \) agents, denoted as \( \mathcal{A} = \{1, 2, \dots, N\} \), operating on a graph \( G = (V, E) \), where \( V \) represents the set of discrete locations and \( E \) denotes the connectivity between them. 
\yn{In this paper, we consider a two-dimensional grid map $G$ with obstacles, where each location has up to four neighbors (up, down, left, right), and obstacles are represented as non-traversable locations.}
Each agent \( i \) is assigned an initial location \( s_i \in V \) and a goal location \( g_i \in V \). Time is discretized into timesteps, and at each timestep, an agent can either move to an adjacent vertex or remain stationary. A solution to MAPF is a set of conflict-free paths \( \Pi = \{\pi_1, \pi_2, \dots, \pi_K\} \) connect agents from initial locations to goals, where each path \( \pi_i \) is a sequence of locations traversed by agent \(i \). A solution is feasible if it avoids:
\begin{itemize}
    \item \yn{\textbf{Obstacle Conflicts}: No agent occupies an obstacle vertex at any timestep.}
    \item \textbf{Vertex Conflicts}: No two agents occupy the same vertex at the same timestep.
    \item \textbf{Edge Conflicts}: No two agents traverse the same edge in opposite directions at the same timestep.
\end{itemize}
\end{definition}

Note that MAPF solutions typically assume agents move synchronously in discrete timesteps, which may differ from real-world robotic systems that involve continuous physical dynamics. Our environment follows such standard MAPF assumption of discrete timesteps in the literature, and considers five movement types which are \textit{up}, \textit{down}, \textit{left}, \textit{right}, and \textit{wait}. The objective of one-shot MAPF is to minimize total travel time
(flowtime) or the completion time of the last agent (makespan). While MAPF has been extensively studied, it features a static one-shot scenario where each agent is assigned a single pre-defined goal and the task concludes upon reaching the goal. This limits applicability in real-world settings where agents must continuously perform new tasks, \yn{motivating the lifelong MAPF setting.}

\begin{definition}[Lifelong MAPF]  
\textit{Lifelong Multi-Agent Path Finding (Lifelong MAPF)} extends MAPF (Definition 1) to dynamic environments where agents are continuously assigned new goal locations upon completing their previous tasks, \yn{reflecting ongoing operations in real-world environments}. Formally, given a set of agents \( \mathcal{A} \) and a graph \( G = (V, E) \), each agent \( a_i \) is \hz{assigned \yn{an infinite} sequence of tasks}, where each task consists of reaching a dynamically assigned goal location \( g_i^t \in V \) at time \( t \).
\end{definition}

Unlike standard MAPF, which terminates when all agents reach their initial goals, lifelong MAPF operates in an online setting, where a task assigner dynamically provides new goal locations throughout the operation. The objective of lifelong MAPF is to maximize \textbf{throughput}, defined as the total number of agents successfully reaching their goal locations over a given time horizon \( T \), while ensuring that paths remain conflict-free. Lifelong MAPF is particularly relevant for warehouse automation and multi-robot systems, where agents must perform continuous tasks efficiently. \yn{In this paper, we study the challenging lifelong MAPF scenarios.}

\section{Method}
\label{sec:Method}
In this section, we first review prioritized planning, then introduce Rolling-Horizon Prioritized Planning (RH-PP), and finally describe our reinforcement learning guided RL-RH-PP framework.

\subsection{Preliminary}
\label{sub_sec_PP}
Prioritized Planning (PP) \citep{erdmann1987multiple} is a widely adopted decoupled approach for solving MAPF problems, known for its computational efficiency. \yn{Instead of planning paths for all agents simultaneously, PP plans the path for each agent according to a predefined priority order obtained via heuristic schemes and plans their paths sequentially via a low-level single-agent path solver e.g., SIPP \citep{phillips2011sipp}}. Specifically, the path for the highest-priority agent is computed first, considering only static obstacles in the environment. The path for each subsequent agent is then planned by treating the paths of previously prioritized agents as dynamic obstacles, ensuring conflict-free navigation.

\begin{definition}[Priority Order]
A \textit{priority order} is a strict partial order defined over a set of agents \( \mathcal{A} = \{1, 2, \dots, N\} \), represented by a binary relation \( \prec \) on \( \mathcal{A} \). The relation \( i \prec j \) indicates that agent \( i \) has a higher priority than agent \( j \) in the planning sequence. A priority order satisfies the \textbf{transitivity property}: if \( i \prec j \) and \( j \prec k \), then \( i \prec k \). In general, a priority order is a \textbf{partial order}, meaning some pairs of agents may not be comparable.
\end{definition}

\begin{definition}[Total Priority Order]
A \textit{total priority order} is a special case of a priority order where every pair of agents is comparable. Formally, a total priority order \( \prec \) on \( \mathcal{A} \) satisfies the \textbf{totality property}: for any two agents \( i, j \in \mathcal{A} \), exactly one of the following holds: \( i \prec j \) or \( j \prec i \). Under a total priority order, all agents are assigned unique, strict rankings, determining an unambiguous planning sequence.
\end{definition}
For simplicity, we represent a total priority order using a tuple notation. For example, a possible total priority order for three agents could be \( (1 \prec 2 \prec 3) \), meaning agent \( a_1 \) plans first, followed by \( a_2 \), and then \( a_3 \). \hzzz{Traditional PP methods typically consider a total priority order, which is established prior to the planning process. Once the order is defined, planning for each agent is sequentially executed according to this strict ordering \citep{erdmann1987multiple,Bennewitz2002,Berg2005}. In contrast, partial priority orders arise in approaches like PBS \citep{ma2019searching}, where priorities emerge dynamically during the planning process and cannot be predetermined.} In this paper, we only focus on using RL to determine total priority orders for PP, and any reference to a priority order should be understood as a total priority order.

\subsection{Rolling Horizon Prioritized Planning}
\label{subsec:RHPP}
\RABC{In this subsection, we present the motivation, design, and discussion of Rolling Horizon Prioritized Planning (RH-PP), which serves as the backbone for the RL–guided framework introduced in the next subsection.}

\subsubsection{Why Prioritized Planning?}
\RABC{Compared to other MAPF solvers, prioritized planning stands out for its simplicity and efficiency. In contrast, search-based MAPF solvers such as Priority-Based Search (PBS) \citep{ma2019searching} and Conflict-Based Search (CBS) \citep{sharon2015conflict} must maintain an explicit search tree, where each node represents a different priority assignment or conflict resolution step. Such branching often grows exponentially with the number of agents in the worst case. Prioritized planning, on the other hand, bypasses explicit search trees by assigning a fixed priority order to agents and solving their paths sequentially. This results in a computational complexity that scales linearly with the number of agents, making PP a potentially desirable approach for large-scale planning and real-time applications.}

\RB{While prioritized planning may not outperform search-based solvers in one-shot MAPF settings, its simplicity and scalability make it well-suited for lifelong MAPF and learning-guided optimization frameworks. In this paper, we exploit such properties by integrating prioritized planning into a reinforcement learning-guided framework that learns to automatically generate effective priority orders in a data-driven manner. This novel hybrid approach enables the adaptive coordination of agents in dynamic, real-world environments and demonstrates superiority over search-based methods in lifelong MAPF.}

\RB{We also note that PIBT \citep{PIBT} provides a decentralized alternative, where priorities are reassigned locally at each timestep. While this yields excellent scalability, the purely local and greedy nature of PIBT’s decisions often leads to congestion or suboptimal routes in dense environments. By contrast, our RL-RH-PP (formally introduced in Section~\ref{subsubsec:RHPP}) framework leverages rolling-horizon prioritized planning with full priority orders sampled and evaluated per replan, allowing the RL policy to internalize global, multi-step interactions within the horizon while keeping the search space tractable. This balance between scalability and solution quality underpins our choice of prioritized planning as the backbone for RL-guided solver for lifelong MAPF.}

\subsubsection{Rolling Horizon Prioritized Planning}
\label{subsubsec:RHPP}

We now introduce the planning backbone Rolling-Horizon Prioritized Planning (RH-PP). RH-PP extends Prioritized Planning (PP) into a dynamic, lifelong setting by iteratively replanning paths in response to newly assigned tasks. Despite the long-standing perception of PP as a simple and suboptimal heuristic for one-shot MAPF, we observe that RH-PP, when combined with randomly sampled total priority orders, performs surprisingly well in lifelong MAPF scenarios. 

Building on the concept of Rolling-Horizon Conflict Resolution (RHCR) \citep{RHCR}, RH-PP divides both path planning and execution into discrete episodes that unfold sequentially over time.  \yn{Each episode is governed by two key parameters: the planning horizon $w$ and the execution horizon $h$, where we must have $h \leq w$.} In each episode, the solver replans paths using PP within a bounded planning horizon \( w \), following a total priority order \( \prec \). Based on such priority order, the low-level single-agent pathfinder SIPP \citep{phillips2011sipp} computes paths to visit a sequence of goal locations known at the current timestep, while treating the paths of already-planned higher-priority agents as dynamic obstacles. \yn{Once all paths are obtained, agents execute their movements for the first \( h \) timesteps.} After completing the execution horizon, a new planning cycle begins, repeating the process with planning horizon \( w \) for solving the lifelong MAPF.

A key challenge in PP lies in its reliance on a fixed total priority order, which strongly affects solution quality. Poor prioritization can cause congestion and unnecessary detours, particularly in dense, constrained environments \citep{Berg2005}. In RH‑PP, determining an effective priority order at each planning step is even more critical, as decisions made within the current horizon directly impact the feasibility and efficiency of future planning. Since RH‑PP operates in a rolling-horizon manner, suboptimal priority assignments can cause early commitments that lead to long-term conflicts or deadlocks, complicating the search for high-quality solutions in later episodes.

Traditional one-shot PP-based solvers primarily focus on \yn{a search process} for high-quality priority assignments for agents. In \cite{Azarm1996}, all possible priority assignments are evaluated before execution. \cite{Bennewitz2002} employs search with hill-climbing to identify high-quality orders, while \cite{Berg2005} proposes a distance query heuristic to construct a good priority order. However, in high-obstacle-density environments, finding a high-quality priority order can be particularly challenging. This difficulty stems from the combinatorial nature of priority ordering, where the number of possible assignments grows exponentially in complex MAPF instances. 

To address these challenges, we first propose a novel yet intuitive heuristic approach that has not been explored in the literature \yn{to our knowledge}. RH-PP starts with a potentially infeasible priority order obtained through top-$K$ random sampling guided by heuristics. We define the heuristic cost function for evaluating each candidate priority order \( \prec_k \) as follows:

\begin{equation}
\label{eq: cost}
    \text{cost}(\prec_k) = - \frac{1}{N}\Big(\sum_{i=1}^{N} e_i + \beta s_i \Big),
\end{equation}
where:
\begin{itemize}
    \item \( e_i \) is the path length of agent \( i \) in the initial solution computed under priority order \( \prec_k \). If agent \( i \) fails to find a feasible path using the single-agent planner, then we make agent $i$ follow its shortest path, and \( e_i \) is the length of the shortest path.
    \item \( s_i = 1 \) if agent \( i \) follows its shortest path, making the initial solution infeasible; otherwise, \( s_i = 0 \).
    \item \( \beta \) is a trade-off that heavily penalizes infeasible partial solutions.
\end{itemize}

The priority order with the lowest cost, $\prec^*$, is selected for execution. This heuristic favors priority orders that initially yield more feasible paths, i.e., those requiring fewer agent repairs, thereby promoting more efficient overall planning. The parameter \( K \) serves as a trade-off between solution quality and efficiency. A larger $K$ increases the number of priority orders evaluated, potentially leading to a better solution but at a higher computational cost. Conversely, a smaller $K$ improves efficiency but may yield less optimal paths.

Note that prioritized planning with any given total priority order can be incomplete  \citep{ma2019searching,stern2019}. A fundamental requirement in RH-PP is ensuring that the final selected order from the Top-K set produces collision-free paths within the planning horizon \( w \). \FN{To achieve this, we adopt a repair mechanism from RHCR \citep{RHCR}, which converts an initially infeasible set of per-agent intended moves into a feasible conflict-free movement within the $w$ window by iteratively resolving local conflicts.} Specifically, the repair process iteratively applies local conflict resolution at each timestep: If the initial solution is infeasible, agents are processed in random order, their desired moves are checked for conflicts, and necessary wait actions are inserted to prevent collisions. \FN{This mechanism guarantees a conflict-free schedule within the $w$-step horizon, but it is purely a safety repair: it does not guarantee progress, completeness, or deadlock resolution. For example, in cyclic blocking configurations (e.g., a loop where each agent wants to move into the next agent’s cell), the repaired execution may stall or exhibit livelock/looping while remaining collision-free. Empirically, such deadlock behaviors are uncommon in our evaluated scenarios for RH-PP. We use the repair only to maintain safe execution under rolling-horizon planning, not to resolve all deadlocks. }

\subsection{Reinforcement Learning-Guided Rolling Horizon Prioritized Planning}

\yn{As established, }the quality of solutions generated by RH-PP is highly dependent on the choice of the total priority order.  \yn{In the previous section, we introduced a simple yet intuitive heuristic for generating such orders. However, in lifelong MAPF, relying on randomly sampled orders is often insufficient for finding effective priority orders at each planning step, as decisions made in the current step directly influence future planning outcomes.} \RC{This continual re-planning setting is fundamentally different from one-shot MAPF: new tasks continually arrive, congestion patterns evolve as agents execute sequences of goals, and short-sighted decisions can create cascading bottlenecks or deadlocks later.} Capturing these long-term dependencies requires a more adaptive and forward-looking strategy. 

Addressing the above challenges naturally aligns with a sequential decision-making process, which makes reinforcement learning (RL) well-suited for generating such orders. \yn{Meanwhile, RH-PP provides a structured yet flexible backbone for integrating RL: 1) its rolling-horizon framework ensures that planning decisions are dynamically revisited and refined; and 2) RL policy can generate promising priority orders in a data-driven manner, substantially reducing the search space for finding the best priorities.} \RC{This combination directly addresses the lifelong MAPF challenges: the rolling horizon allows continual adaptation to new goals, while RL-guided sampling mitigates cascading inefficiencies by reasoning over long-horizon effects.} In this section, we introduce an RL approach that intelligently generates priority orders dynamically at each timestep, optimizing for both immediate feasibility and long-term throughput. 

\begin{figure*}[t!]
\centering
\includegraphics[width=0.7\textwidth]{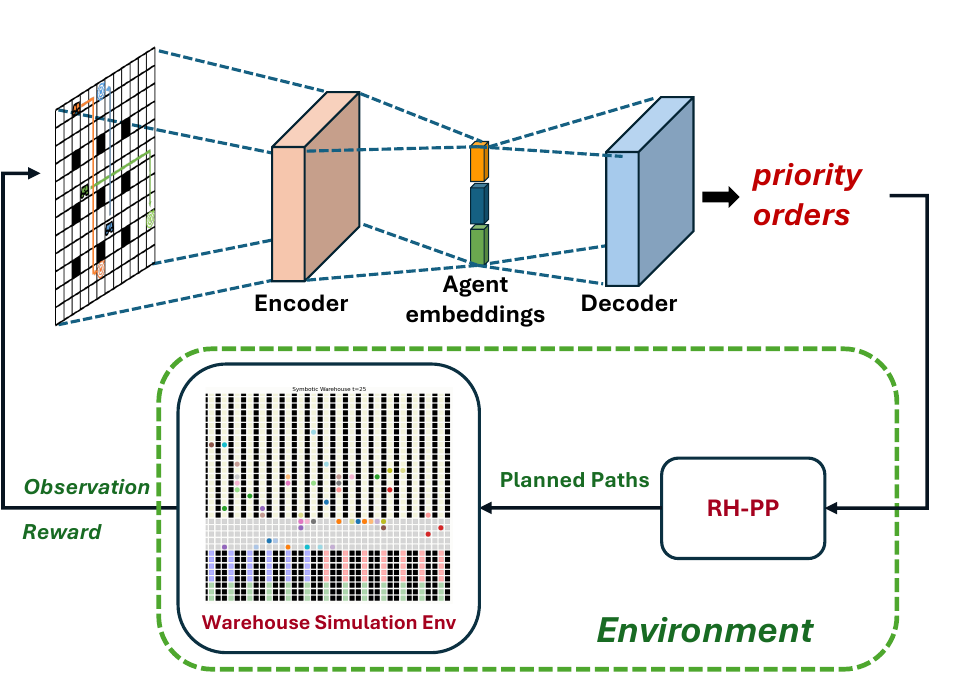}
\caption{\RC{The framework of our proposed RL-RH-PP.}
At each planning step, the RL policy encodes \RC{shortest path of each agents} into agent embeddings and autoregressively decodes a total of $K$ priority orders. These orders are passed to the RH-PP planner to compute the conflict\yn{-free} path, which is then executed in the warehouse simulation. Feedback rewards are used to update the policy. This closed-loop interaction enables the RL policy to learn to generate effective priority orders during training.}
\label{fig:1}
\end{figure*}

To this end, we first reformulate the task of selecting an optimal total priority order at each planning step as a \hz{Partially Observable Markov Decision Process (POMDP)}. We then propose RL-RH-PP, a reinforcement learning-guided rolling horizon prioritized planning framework.
The overall framework is illustrated in Figure~\ref{fig:1}. The RL policy interacts with an environment that includes a warehouse simulation and an RH-PP planner. At each planning step, the RL policy generates $K$ total priority orders, those orders are passed to RH-PP and evaluated according to Eq~\ref{eq: cost}, the selected $\prec^*$ is passed to subsequent planning to compute feasible paths for the agents. These paths are then executed within the warehouse simulation. Feedback rewards from the executed paths and next observations are used to update the RL policy. This iterative learning process enables the policy to refine its decision-making on sampling priority orders, progressively improving the quality of the total priority order and the resulting agent paths over multiple training episodes.

\RC{Our RL formulation is particularly beneficial in this lifelong setting, where sequential decision-making and adapting to new goals are essential to prevent cascading congestion, but these characteristics may not reflected in one-shot MAPF.}

\subsubsection{POMDP Formulation}
\label{subsubsction: POMDP fomulation}

We define the \hzz{POMDP considered in this paper as follows}, where:

\begin{itemize}
\item 
\textbf{State}: \hzzz{The state should encompass all agent locations, their goal lists, and any environment details needed to maintain the Markov property. However, explicitly modeling all such details can be cumbersome. Instead, we rely on partial observations-the agents’ shortest paths-to capture the key spatiotemporal information for conflict resolution. The underlying transitions remain governed by the full (but unmodeled) state and the RH-PP module, while the agent learns directly from these compact observations.}
\item \textbf{\hzz{Observation}}: The observation $\boldsymbol{o}_t$ at time step $t$ encapsulates essential information for the RL to determine the effective priority order. In this paper, we consider the observation to be \hz{the shortest paths $\sigma_{i}^t$ that connect the current location of an agent $i$ at the planning time step $t$ to its future goal sequence. Mathematically, we have $\boldsymbol{o}_t = [\sigma_{1}^t,\sigma_{2}^t,...\sigma_{N}^t]$. Note we define the position on the MAPF map using integers, then we have $\sigma_{i}^t \in \mathbb{Z}^r$, where $r$ is the (maximum) length of the shortest path.}
\hzz{Shortest paths are a natural and informative choice for representing MAPF observations in the literature, e.g., \cite{yan2024neural}, as they capture both spatial and temporal information about an agent's intended trajectory. By including the full sequence of planned steps toward a goal, shortest paths implicitly encode potential conflicts, path overlaps, and estimated completion times, all of which are relevant for downstream prioritization.}

\item \textbf{Action}: For a MAPF instance with $N$ agents, the space of all possible total priority orders is $N!$. \yn{We consider the action to be a set of $K$ promising priority orders, which is used to reduce the search space to guide the RH-PP heuristics.} Those orders are then passed to RH-PP to obtain the final feasible path.

\item \textbf{Reward function}: \RALL{The reward $R$ guides the learning process and is defined as:
\begin{equation}
\label{eq:rl_reward}
    R(\boldsymbol{o}_t, a_t) = - \frac{1}{N} \Big( \sum_{i=1}^{N} d_{i,t} + \kappa c_{i,t}  + \sigma s_{i,t}\Big).
\end{equation}}
 
\RC{In the lifelong setting, at time $t$ agent $i$ may have multiple outstanding goals arranged in a sequence. We define $d_{i,t}$ as the average Manhattan distance from the agent’s current location to each of these future goal locations. }\RALL{The congestion penalty $c_{i,t}$ is a binary indicator equal to 1 at time $t$ if the agent’s planned actions for the next $h$ steps are all wait (the agent is locally stalled), and 0 otherwise. The infeasibility penalty $s_{i,t}$ is a binary indicator equal to 1 at time $t$ if, under the selected priority order, RH-PP fails to find a feasible path for agent $i$, and 0 otherwise. \FN{To keep the binary penalties on a scale comparable to the distance term, we introduce weighting factors $\kappa$ and $\sigma$, both set empirically to $1000$. As shown in our experiments (Section~\ref{subsubsec:pentaly weights}), smaller weights slow convergence, while extremely large weights provide little additional benefit. Note that these weights are intended to match the magnitude of the total distance summed over all agents; therefore, they should scale with the map size (e.g., warehouse width + length), since larger maps typically induce longer average path lengths.}
 Maximizing the return under this reward encourages fast progress while actively avoiding congestion and infeasible orders, thereby improving overall throughput.}

\hzzz{Note that our reward function mirrors the heuristic cost in Eq.\ref{eq: cost} but differs in a crucial way. Whereas $e_i$ in Eq.\ref{eq: cost} is the prospective pre-execution path length computed at planning time, $d_{i,t}$ measures the actual remaining distance after the agent has already moved. Consequently, $d_{i,t}$ more directly reflects each agent’s real progress and is thus more informative for increasing overall system throughput.}

\item \textbf{Discount factor}: The discount factor $\gamma=0.99$ determines the importance of future rewards, promoting long-term planning.

\end{itemize}

\hzz{Note that the transition function of our POMDP is implicitly determined by RH-PP}. The RL policy learns directly from data collected by interacting with the warehouse simulation environment and observing state transitions \yn{on the fly during training}.

\subsection{Neural Architecture Design for RL-RH-PP}

To effectively generate total priority orders that guide RH-PP, we employ a neural network composed of an \textit{encoder} and an autoregressive \textit{decoder}. The encoder extracts informative embeddings for each agent from the environment observations, while the decoder autoregressively generates the total priority orders.

\subsubsection{Encoder Design}
The encoder is responsible for transforming the raw input observation into a compact representation that effectively captures both spatial and temporal information about agents and obstacles in the environment. Given the observation at time step \(t\), defined as $
\boldsymbol{o}_t = \bigl[\sigma_{1}^t, \sigma_{2}^t, \dots, \sigma_{N}^t\bigr]
$ in section~\ref{subsubsction: POMDP fomulation},
\hzz{we employ attention-based encoders} to extract agent embeddings.

\begin{figure*}[t!]
\centering
\includegraphics[width=\textwidth]{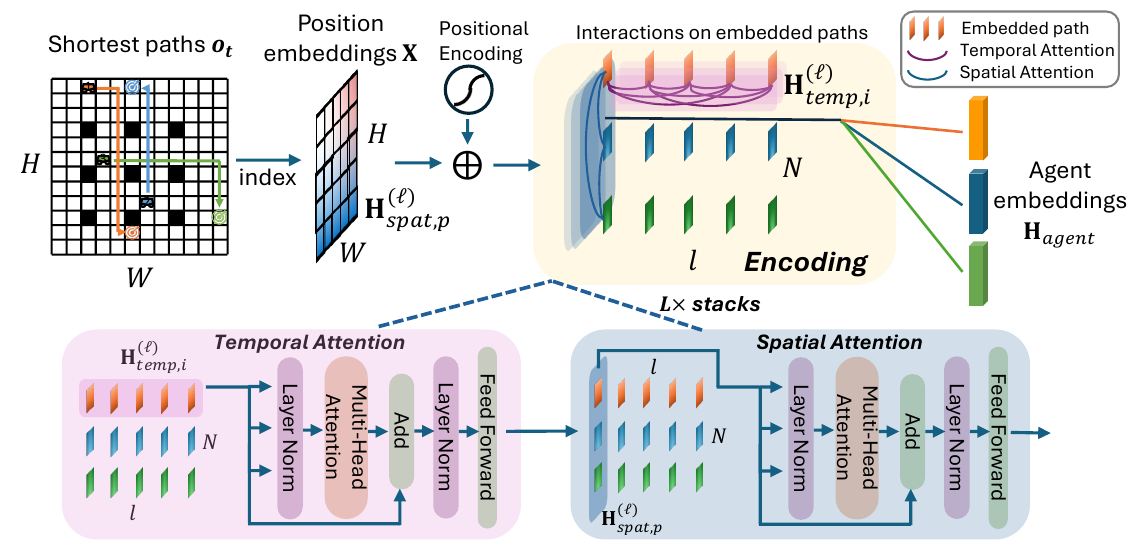}
\caption{\RC{Our proposed encoding process for RL-RH-PP. Raw path inputs are indexed using position embeddings to construct a learnable representation for agent paths. Temporal and spatial attention mechanisms are applied in turn, along with normalization and residual modules, to extract and refine agent embeddings.}}
\label{fig:encoder}
\end{figure*}

\yn{Figure~\ref{fig:encoder} depicts the encoding process.} To effectively represent the observation, we introduce a dictionary of learnable position embeddings, \(\mathbf{X} \in \mathbb{R}^{d \times W \cdot H}\). Each column of \(\mathbf{X}\) corresponds to a drivable location in the map, where \(d\) denotes the dimension of each embedding and \(W\) and \(H\) are the width and height of the warehouse map. \RB{This representation design provides generalization capability, enabling zero-shot transfer to environments with different numbers of agents and planning horizons, as well as to unseen maps with varied topologies (e.g., obstacle rearrangements or corridor geometries), while holding map size fixed.}

For each agent \(i\), its shortest path \(\sigma_{i}^t\) is used to index the embedding dictionary. Specifically, for each integral location \(\sigma_{i,m}^t\) at index $m$ in \(\sigma_{i}^t\), we retrieve the corresponding column \(\mathbf{X}[\sigma_{i,m}^t]\). This results in a sequence of embeddings that represent the agent's raw path. Formally, the representation \(h_i^t\) of the input path is defined as:
\begin{equation}
    \mathbf{h}_{i}^t 
    = 
    \hz{\bigl[\mathbf{X}[\sigma_{i,0}^t],\, \mathbf{X}[\sigma_{i,1}^t],\, \dots,\, \mathbf{X}[\sigma_{i,r}^t]\bigr] \in \mathbb{R}^{r \times d}.}
\end{equation}

We add sinusoidal positional encoding to every \(h_i^t\) before feeding them to temporal and spatial encoding layers following the Transformer design \citep{vaswani2017attention}.

\paragraph{Temporal and Spatial Encoding.}
To encode both the temporal progression of each agent’s trajectory and the spatial relationships among agents, we stack \(L\) encoder layers. Each layer \(\ell\) applies two multi-head self-attention (MHA) blocks sequentially: one for temporal attention and one for spatial attention. In addition, we adopt pre-layer normalization and include a residual connection \citep{he2016deep} within each attention block. This design integrates temporal information from each agent’s trajectory and spatial interactions among all agents, while the residual connections and layer normalization \citep{ba2016layer} stabilize training and enable deeper stacking of layers. Let \(\mathbf{H}^{(\ell-1)}\in \mathbb{R}^{N \times r \times d}\) be the output of the \((\ell - 1)\)-th layer, \(\mathbf{H}^{(0)} \) as the input to the first layer of the encoder, where \(N\) is the number of agents, \(r\) is the (maximum) shortest path length, and \(d\) is the embedding dimension. \hz{In the rest within this section, we refer \textit{agent dimension} as indexing a tensor along its $N$ dimension while \textit{time dimension} along its $r$ dimension.} The \(\ell\)-th layer proceeds as follows.

\paragraph{Temporal Attention.} 
    
    Let \(\mathbf{H}^{(\ell-1)}_{\text{temp},i} \in \mathbb{R}^{r \times d}\) be the index of \(\mathbf{H}^{(\ell-1)}\) along agent dimension for agent $i$, We first normalize:
    \begin{equation}
        \widetilde{\mathbf{H}}^{(\ell)}_{\text{temp},i}
        = \mathrm{LayerNorm}\bigl(\mathbf{H}^{(\ell-1)}_{\text{temp},i}\bigr).
    \end{equation}
    Then we feed \(\widetilde{\mathbf{H}}^{(\ell)}_{\text{temp},i}\) into a multi-head self-attention module operating along the time dimension $l$ for each agent independently. We project \(\widetilde{\mathbf{H}}_{\text{temp},i}\) into queries, keys, and values:
    \begin{equation}
        \mathbf{Q} 
        = 
        \widetilde{\mathbf{H}}^{(\ell)}_{\text{temp},i} \,\mathbf{W}^{Q}, 
        \quad 
        \mathbf{K} 
        = 
        \widetilde{\mathbf{H}}^{(\ell)}_{\text{temp},i} \,\mathbf{W}^{K}, 
        \quad 
        \mathbf{V} 
        = 
        \widetilde{\mathbf{H}}^{(\ell)}_{\text{temp},i} \,\mathbf{W}^{V},
    \end{equation}
    and compute attention in each head \(u\) using
    \begin{equation}
        \mathrm{head}_u 
        = 
        \mathrm{Softmax}\!\Bigl(
            \tfrac{\mathbf{Q}_u \mathbf{K}_u^\top}{\sqrt{d/U}}
        \Bigr)\,\mathbf{V}_u,
    \end{equation}
    \hz{where \(U\) is the number of heads and \(\mathbf{Q}_u, \mathbf{K}_u, \mathbf{V}_u\) denote the \(u\)-th partitions of \(\mathbf{Q},\mathbf{K},\mathbf{V}\).} After concatenating all heads and applying an output projection \(\mathbf{W}^O\), we obtain:
    \begin{equation}
        \widehat{\mathbf{H}}^{(\ell)}_{\text{temp},i} 
        = 
        \mathrm{Concat}\bigl(\mathrm{head}_1,\dots,\mathrm{head}_H\bigr)\,\mathbf{W}^O,
    \end{equation}
    following a residual connection:
    \begin{equation}
        \mathbf{H}^{(\ell)}_{\text{temp},i} 
        = 
        \widetilde{\mathbf{H}}^{(\ell)}_{\text{temp},i}
        + 
        \widehat{\mathbf{H}}^{(\ell)}_{\text{temp},i}.
    \end{equation}

\paragraph{Spatial Attention.}
    We applied the temporal attention for every agent, we stack back the output to get \(\mathbf{H}^{(\ell)}_{\text{temp}} \in \mathbb{R}^{N \times r \times d}\), \hz{and index it along the time dimension for every time step \(p\)} to get \(\mathbf{H}^{(\ell)}_{\text{spat}, p} \in \mathbb{R}^{N \times d}\). We then normalize:
    \begin{equation}
        \widetilde{\mathbf{H}}^{(\ell)}_{\text{spat},p}
        = \mathrm{LayerNorm}\bigl(\mathbf{H}^{(\ell)}_{\text{spat},p}\bigr),
    \end{equation}
    and apply multi-head self-attention across the agent dimension $N$ for each time step, reusing the same multi-head attention (MHA) formulation but focusing on agent-to-agent interactions:
    \begin{equation}
        \widehat{\mathbf{H}}^{(\ell)}_{\text{spat},p} 
        = 
        \mathrm{MHA}\bigl(\widetilde{\mathbf{H}}^{(\ell)}_{\text{spat},p}\bigr).
    \end{equation}
    One can interpret this attention mechanism as operating on an agent-agent graph \citep{graphomer} at each time step, thereby capturing global interactions among all agents. The residual connection here yields:
    \begin{equation}
        \mathbf{H}^{(\ell)}_{\text{spat},p} 
        = 
        \widetilde{\mathbf{H}}^{(\ell)}_{\text{spat},p} +
        \widehat{\mathbf{H}}^{(\ell)}_{\text{spat},p} 
    \end{equation}

    We finally stack $\mathbf{H}^{(\ell)}_{\text{spat},p}$ for every time step $p$ to obtain the output from the encoder \(\mathbf{H}^{(\ell)}\in \mathbb{R}^{N \times r \times d}\) .

Following each attention block, we apply the feed-forward layers with ReLU activation following the original Transformer \citep{vaswani2017attention}. These layers use the same logic on layer normalization and a residual connection as the attention layer, maintaining stable gradient flow and facilitating deeper stacking. This additional transformation further refines agent representations before passing them to the next encoder layer.

\paragraph{Agent Embeddings.} Repeating this process for \(\ell = 1, 2, \ldots, L\) produces a final representation 
\(\mathbf{H}^{(L)}\in \mathbb{R}^{N \times r \times d}\). 
We then choose the embeddings at the first index along the time dimension of \(\mathbf{H}^{(L)}\) as the agent embeddings, denoted as \(\mathbf{H}_{agent} \in \mathbb{R}^{N \times d}\). And \(\mathbf{H}_{agent,i} \in \mathbb{R}^{d}\) is the final agent-level feature for agent \(i\).

\paragraph{\yn{Remarks.}} The encoder is designed to capture both temporal and spatial dependencies efficiently. Temporal attention enables each agent to aggregate information from its trajectory, while spatial attention models interactions among agents at each time step. By stacking multiple layers, the model can learn hierarchical representations, capturing complex dependencies that are critical for effective multi-agent coordination.

\begin{figure*}[t!]
\centering
\includegraphics[width=\textwidth]{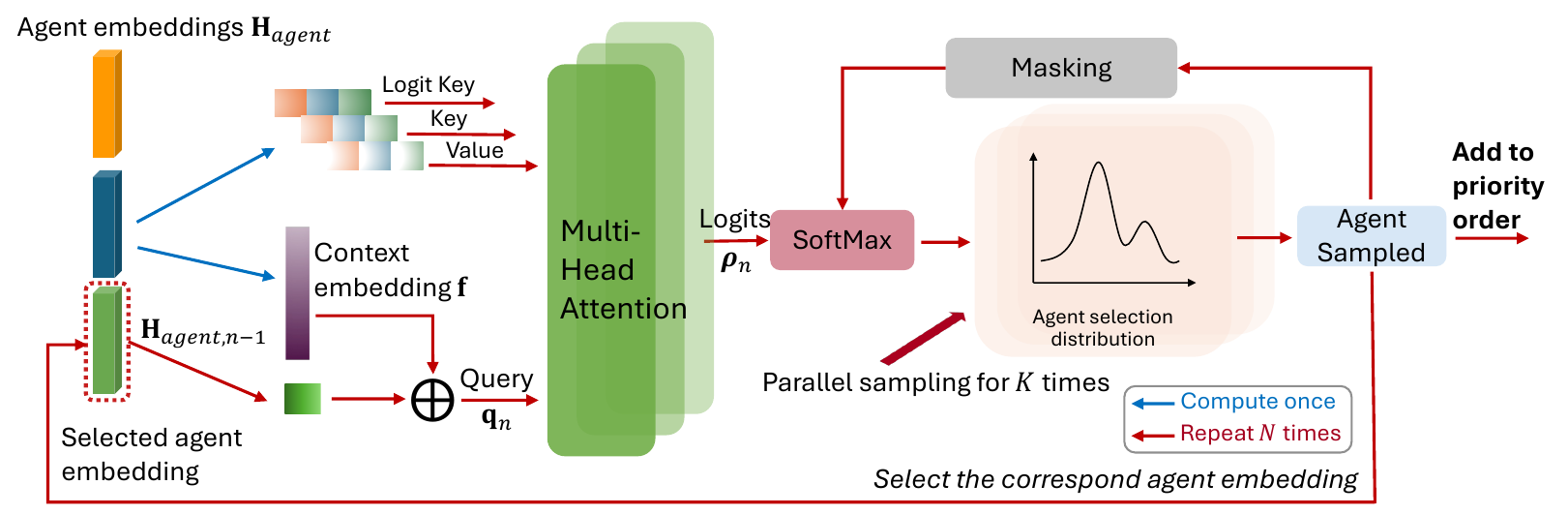}
\caption{\hz{Our proposed decoding process for RL-RH-PP. Agent embeddings from the encoder are fed into a self-attention mechanism to generate a sampling distribution. To construct each priority order, agents are sampled and added one by one in an autoregressive manner until all $N$ agents are assigned. This process can be executed in parallel to generate a set of $K$ promising orders efficiently.}}
\label{fig:decoder}
\end{figure*}

\subsubsection{Decoder Design}
After obtaining the agent embeddings \(\mathbf{H}_{agent} \in \mathbb{R}^{N \times d}\) from the encoder for \(N\) agents, the decoder generates total priority orders in an autoregressive manner. \yn{Figure~\ref{fig:decoder} depicts the decoding process. Specifically, at each decoding step, one agent is selected until all $N$ agents are chosen exactly once, forming a priority order (permutation) of the $N$ agents. Finding the optimal priority order over all $N$ agents constitutes a combinatorial optimization problem. Our decoder is inspired by recent advances in neural combinatorial optimization, particularly the attention model \citep{kool2018attention} originally developed for vehicle routing problems.}

\paragraph{Autoregressive Decoding with  Multi-Head Attention.}
We first project \(\mathbf{H}_{agent}\) into keys, values, and glimpse keys:
\begin{equation}
\mathbf{K} = \mathbf{H}_{agent}\,\mathbf{W}^{K}, 
\quad
\mathbf{V} = \mathbf{H}_{agent}\,\mathbf{W}^{V}, 
\quad
\mathbf{L} = \mathbf{H}_{agent}\,\mathbf{W}^{L},
\label{eq:fixed-proj}
\end{equation}
where \(\mathbf{W}^K, \mathbf{W}^V, \mathbf{W}^L \in \mathbb{R}^{d \times d}\) are learnable projection weights. We also compute a global embedding by averaging \(\mathbf{H}_{agent}\) in the agent dimension and project it to a fixed context \(\mathbf{f} \in \mathbb{R}^d\). These quantities in \eqref{eq:fixed-proj} and \(\mathbf{f}\) are only computed once and remain unchanged throughput decoding to enhance efficiency.

\yn{In the following section, we describe our decoder and how it generates a single priority order. In practice, $K$ orders can be efficiently generated in parallel, enabled by batched autoregressive decoding in deep learning frameworks such as PyTorch.}
The decoding proceeds in \(N\) steps, where at each decoding step $n$ we sample one agent and then add to the total priority order. At each step \(n\), we form a decoding step context \(\mathbf{q}_n\) by combining the fixed context \(\mathbf{f}\) with the embedding of the agent selected at the previous decoding step. If \(n=1\), we use a learned placeholder instead of a previously selected agent. Concretely,
\begin{equation}
\mathbf{q}_n
= 
\mathbf{f} 
\,\,+\,\,
\mathbf{W}^{Q_{\text{step}}}
\bigl(\mathbf{H}_{agent, n-1}\bigr),
\end{equation}
where \(\mathbf{H}_{agent, n-1}\in \mathbb{R}^{d}\) is the embedding of the agent chosen at decoding step \(n-1\) (or the placeholder for \(n=1\)), and 
\(\mathbf{W}^{Q_{\text{step}}} \in \mathbb{R}^{d\times d}\)
is a learnable projection.

\hz{We then broadcast \(\mathbf{q}_n\) across \(U\) heads to form \(\mathbf{q}_{n,u}\). For each head \(u\), we compute a dot-product attention with the keys and values:}
\begin{equation}
\mathrm{head}_u
=
\mathrm{Softmax}
\Bigl(
\tfrac{\mathbf{q}_{n,u} \,\mathbf{K}_u^\mathsf{T}}{\sqrt{d/U}}
\Bigr)
\,\mathbf{V}_u,
\label{eq:head-compute}
\end{equation}
where \(\mathbf{K}_u,\mathbf{V}_u\) are the partitions of \(\mathbf{K}, \mathbf{V}\) for head \(u\). Concatenating the heads along the embedding dimension and applying a linear projection yields a glimpse vector \(\mathbf{g}_n\  \in \mathbb{R}^d\). We compute unnormalized logits for each agent by
\begin{equation}
\boldsymbol{\rho}_n
=
\mathbf{L} \mathbf{g}_n,
\end{equation}
where \(\mathbf{L} \in \mathbb{R}^{N \times d}\) is defined in \eqref{eq:fixed-proj}. 

\paragraph{Sampling and Masking.}
Let \(\mathcal{U}_{n-1}\) be the set of agents already chosen in the first \((n-1)\) decoding steps. We impose a mask over \(\boldsymbol{\rho}_n\) by setting
\(\boldsymbol{\rho}_n[j] = -\infty\)
for every agent \(j\in \mathcal{U}_{n-1}\). The decoder then applies a softmax to the remaining logits:
\begin{equation}
p(a_n = i \,\mid\, \mathbf{q}_n, \mathcal{U}_{n-1})
=
\frac{\exp\bigl(\boldsymbol{\rho}_n[i]\bigr)}{
\sum_{v \notin \mathcal{U}_{n-1}}
\exp\bigl(\boldsymbol{\rho}_n[v]\bigr)}.
\end{equation}
To select an agent \(a_n\) at decoding step \(n\), we sample from this categorical distribution:
\begin{equation}
a_n \sim p(a_n \,\mid\, \mathbf{q}_n, \mathcal{U}_{n-1}),
\end{equation}
thereby ensuring that no agent is chosen more than once.

\paragraph{\hzz{Top-K sampling of Priority Orders.}}
By repeating this procedure for \(n=1,2,\ldots,N\), the decoder generates a permutation of $N$ agents
that includes each agent exactly once. Leveraging batch parallelism, the decoder generates $K$ priority orders simultaneously in parallel.
\hz{These orders are then passed to the RH-PP planner to solve for conflict-free paths. Please refer to Table~\ref{tab:hyperparam_table} in the Appendix for detailed network hyperparameter selections.}

\subsubsection{RL Training}
To train the reinforcement learning policy for generating effective total priority orders, we adopt Proximal Policy Optimization (PPO) \citep{schulman2017proximal}, a widely used on-policy RL algorithm known for its stability and sample efficiency. PPO optimizes the policy by iteratively interacting with the environment, collecting trajectories, and updating the policy while ensuring constrained updates to prevent instability. Algorithm~\ref{alg:RL Training} shows how our training for RL-RH-PP.  We implement an actor-critic PPO framework where the policy and value networks share the same encoder architecture but maintain separate parameters. The value network uses the same encoder architecture as the policy network to obtain agent embeddings, which are then processed by a value decoder consisting of an attention layer followed by linear layers to produce a scalar value, similar to \cite{ma2021learning}.

\begin{algorithm}[t]
\caption{Training for RL-RH-PP}
\label{alg:RL Training}
\begin{algorithmic}[1]
    \State \textbf{Input:} Initial policy parameters $\theta_0$, initial value function parameters $\phi_0$
    \For{$z = 0, 1, 2, \dots, Z$} \hz{\Comment{Policy update iterations}}
        \If{$z \mod f =0$} \hz{\Comment{Start collecting rollouts, reuse for $f$ updates}}
        \For{$t=0,1,2,\dots, T$} \hz{\Comment{Simulate}}
            \If{$t \mod h = 0$} \hz{\Comment{Replan paths}}
                \State \RC{sample $K$ priority orders  from $\pi_{\theta_z}$ and pass them to RH-PP for path planning}
            \EndIf
        \EndFor
        \State Collect set of trajectories $\mathcal{D} = \{\tau_j\}$
        \EndIf
        \State Compute rewards-to-go $\hat{R}_t$ and advantage estimates $\hat{A}_t$ based on the current \Statex \hspace{0.5cm} value function $V_{\phi_z}$.
        \State Update the policy by maximizing the PPO objective:
        \begin{equation}
            \theta_{z+1} = \arg\max_{\theta} \frac{1}{|\mathcal{D}| {T}} \sum_{ {\tau} \in \mathcal{D}} \sum_{t=0}^{ {T}} 
            \left[
            \min \left( 
                \frac{\pi_{\theta}(a_t|\boldsymbol{o}_t)}{\pi_{\theta_z}(a_t|\boldsymbol{o}_t)} \hat{A}(\boldsymbol{o}_t, a_t), \,
                g(\epsilon, \hat{A}(\boldsymbol{o}_t, a_t)) 
            \right)
            + \eta \mathcal{H}(\pi_{\theta}(\cdot|\boldsymbol{o}_t))
            \right]
        \end{equation}

        \State Fit value function by regression on mean-squared error:
        \begin{equation}
            \phi_{z+1} = \arg\min_{\phi} \frac{1}{|\mathcal{D}| {T}} \sum_{ {\tau}\in \mathcal{D}} \sum_{t=0}^{ {T}} 
            \left( V_{\phi}(\boldsymbol{o}_t) - \hat{R}_t \right)^2,
        \end{equation}
    \EndFor
\end{algorithmic}
\end{algorithm}

To enhance sample efficiency during rollouts, we reuse the rollout dataset \( \mathcal{D} \) for \( f \) iterations in both policy and value updates. We add the entropy loss $\mathcal{H}(\pi_{\theta}(\cdot|\boldsymbol{o}_t))$ to encourage exploration and prevent premature policy collapse. For stable training, we incorporate key implementation techniques, including advantage normalization and gradient clipping, as suggested by \cite{ppotricks}.

\section{Experimental Results}
\label{sec:experiments}

We present a comprehensive evaluation of our proposed RL-RH-PP approach across various warehouse environments and agent densities. Our evaluation is structured around the following key research questions:

\RALL{
\begin{itemize}
    \item \textbf{Q1:} What is the effectiveness of RL training in RL-RH-PP for boosting RH-PP (see Section~\ref{subsec: eft_RL})?
    \item \textbf{Q2:} How does RL-RH-PP compare to existing lifelong MAPF solvers in terms of throughput and solve time at inference time (see Section~\ref{sec:benchmark})?
    \item \textbf{Q3:} How does RL-RH-PP generalize to new scenarios with varying agent densities, planning horizons and warehouse layouts (see Section~\ref{subsec:gen_anay})?
    \item \textbf{Q4:} What is the anytime behavior of RL-RH-PP (see Section~\ref{subsec: anytime})? 
\end{itemize}
}

\RALL{To address these questions, we first outline the experimental setups, including the warehouse simulation environments, baseline methods, and evaluation metrics. We then present our key findings on throughput and solve time across two real-world-inspired warehouse layouts. Next, we assess the zero-shot transferability of the trained RL policy to different agent densities, planning horizons, and unseen maps. Finally, we conduct several ablation studies to analyze the effectiveness of various key designs in our method.}

\subsection{Experimental Setups}
We introduce the details of our experimental setups. 

\subsubsection{Simulation Environment}

We develop a lifelong warehouse simulation environment based on open-source MAPF benchmarks \citep{stern2019multi,RHCR}, using RH-PP as the back-end planner. The simulation takes a total priority order as input and outputs the corresponding reward and observation, as defined in Section \ref{subsubsction: POMDP fomulation}. By default, policy is trained using a planning horizon of $w{=}20$, an execution horizon of $h{=}5$, and a total simulation horizon of $T{=}800$ steps. Experiments were conducted on two warehouse maps, with tasks randomly generated and assigned according to the respective warehouse layout. Specifically, we consider:

\begin{figure}[t]
\centering
\includegraphics[width=0.6\textwidth]{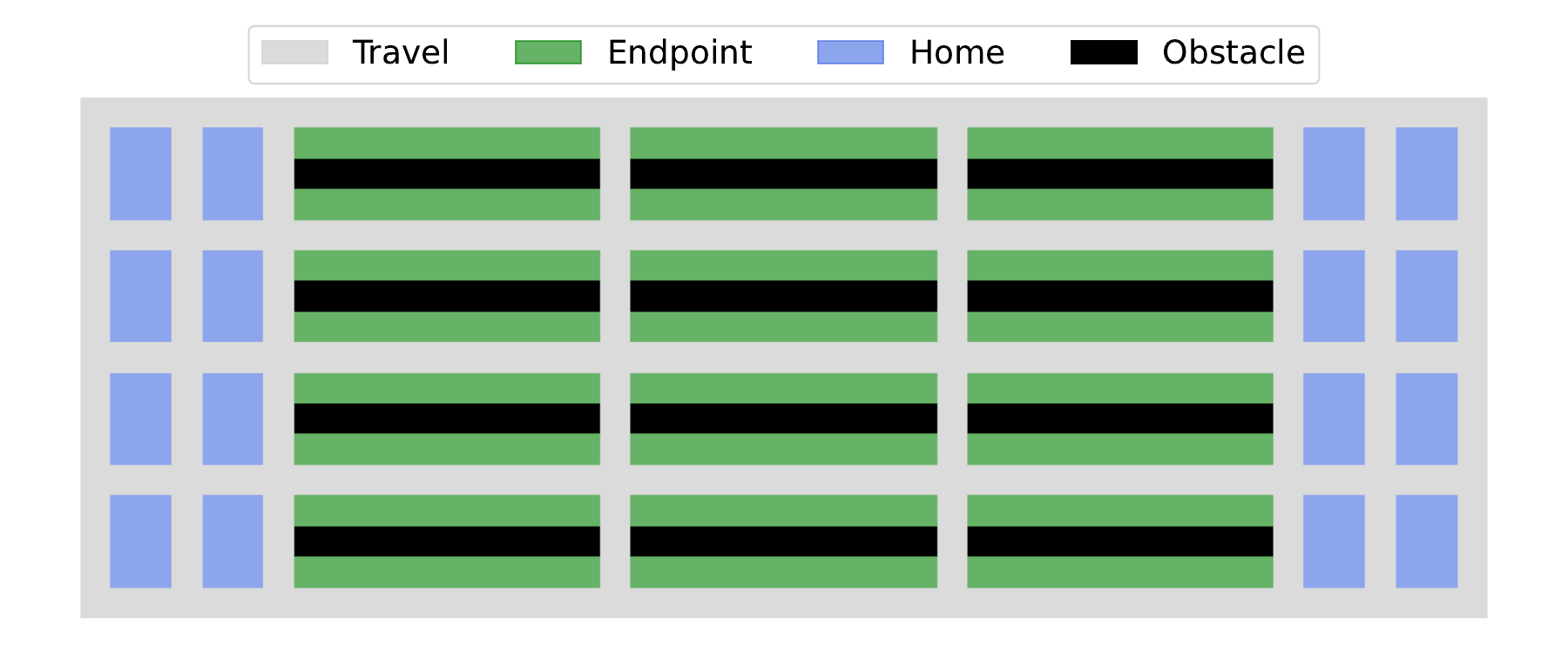}
\caption{Amazon fulfillment center dense map (obstacle density = 15.3$\%$), modified from \citep{li2021lifelong}.}
\label{fig:amazon}
\end{figure}

\begin{figure}[t]
\centering
\includegraphics[width=0.6\textwidth]{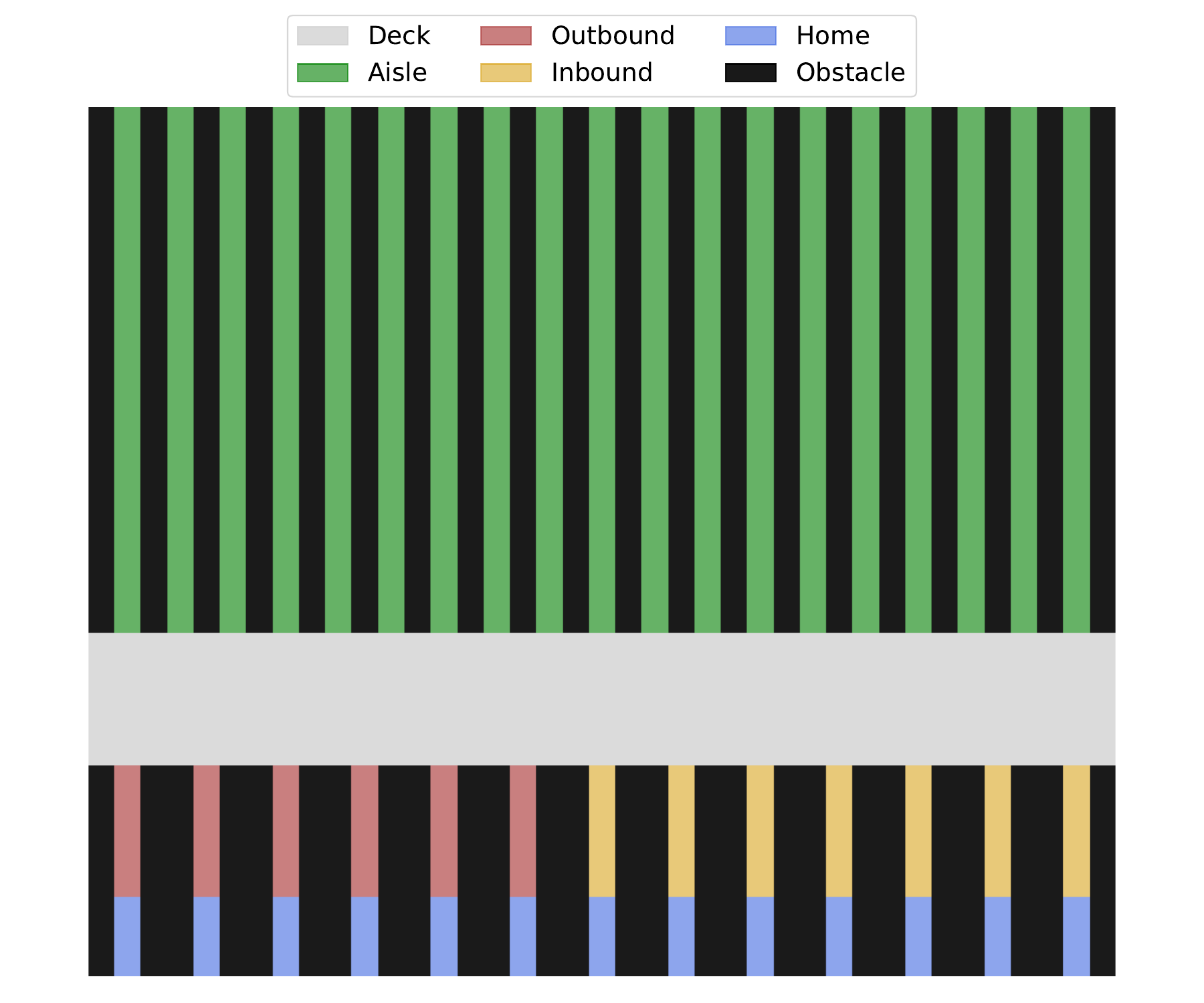}
\caption{Symbotic warehouse map (obstacle density = 56.6$\%$).}
\label{fig:symbotic}
\end{figure}

\begin{itemize}
    \item \textbf{\hzzz{Amazon fulfillment center dense.}} 
    As shown in Figure~\ref{fig:amazon}, this map is characterized by multiple parallel aisles and narrow corridors. To create scenarios with high agent density, we reduce the size of the map to half of its original size in \cite{li2021lifelong}. Agents are randomly initialized in the drivable free space, which consists of home and travel locations, and the goal location candidates are the endpoint locations (green blocks in Figure~\ref{fig:amazon}). When assigning a new task for each agent, the system first checks if the agent’s current goal list is empty. If empty, it randomly selects an endpoint from a predefined set. It ensures that the selected endpoint is not the agent’s current location and the endpoint is not already held by another agent.

    \item \textbf{Symbotic warehouse.}
    As shown in Figure~\ref{fig:symbotic}, this map is already considered a high obstacle density environment, where the layout contains bottlenecks (e.g., small cross-aisles) that can cause agent congestion if not planned properly. Note that the map used in our simulation is shaped like a Symbotic warehouse, but it is not based on exact dimensions. \RC{The entire map is divided into four regions: inbound, outbound, deck, and aisles. Agents handle package operations, where inbound and outbound are used for shipping in and out packages, respectively, and aisles serve as locations for storage and retrieval. The task assignment depends on the last goal location and the agent’s loading state. Initially, all agents are randomly initialized in the deck and inbound areas with packages loaded. If the last goal location is an inbound station, the next destination is randomly assigned as an aisle station, after which the agent transitions to an unloaded state. If the last goal is from an outbound station, the next destination is randomly selected between an inbound station and an aisle station, resulting in the agent becoming loaded. When the last goal is from an aisle station, the next destination depends on the current loading state: an unloaded agent selects randomly between an inbound station and an aisle station and becomes loaded; and a loaded agent is directed to an outbound station, thereby returning to an unloaded state.}
\end{itemize}

In both maps, non-drivable map locations are treated as static obstacles. The Amazon fulfillment center map has an obstacle density of 15.3\%, while the Symbotic warehouse map features a much higher density of 56.6\%. Hence, lifelong MAPF solvers may face significantly greater difficulty in finding feasible paths within the Symbotic map. For each map, we also vary the number of agents to adjust the density level, allowing us to evaluate both performance and scalability in different scenarios. Specifically, we train the RL policy with \( N \in \{80, 100, 120\} \) agents for both warehouse environments. These settings create a high agent density; for instance, in the Amazon fulfillment center, the highest density in our setting is 2 times higher than that reported in \cite{li2021lifelong}.

Finally, to enable more efficient roll-out sampling during training, we parallelize the environment on CPUs using Tianshou \citep{tianshou} and wrap it with OpenAI Gym \citep{opaigym}. These setups ensures scalable and efficient rollout data collection during RL training.

\subsubsection{Training Setups}
We configure the hyperparameters in Algorithm~\ref{alg:RL Training} as follows: the total number of epochs is set to $Z = 4000$, the rollout reuse per epoch to $f = 3$. For the PPO objective, we set the entropy loss weighting factor to $\eta = 0.01$ and the PPO clipping parameter to $\epsilon = 0.2$. The optimization is performed using the Adam optimizer \citep{kingma2014adam} with an initial learning rate of $\lambda = 0.001$ and an exponential decay rate of $0.999$ for both policy and value network updates. We use a batch size $B=32$ and apply gradient clipping with a threshold  $g_{clip} = 0.5$ to stabilize training. All the training hyperparameters are summarized in Table~\ref{tab:hyperparam_table}. 

Training is conducted on a workstation using a single NVIDIA RTX 6000 Ada GPU. For each agent density setting, we train a separate RL policy from scratch. We use $K=1$ for the Top-$K$ sampling in RH-PP to improve rollout efficiency. The longest training duration, corresponding to $N = 120$, is approximately four days.

\subsubsection{Evaluation Setups}
\label{subsec:throughput & time}
To benchmark RL-RH-PP, we compare it against several representative non-learning-based solvers, including Priority-based Search (PBS) \citep{ma2019searching}, Conflict-based Search (CBS) \citep{sharon2015conflict}, \RABC{PIBT \citep{PIBT}} and WPPL \citep{jiang2024scalinglifelongmultiagentpath}. Additionally, we include our proposed method, RH-PP, as part of the baselines.

These solvers are chosen for their relevance and complementary strengths in lifelong MAPF. CBS ensures optimality but struggles with scalability \citep{sharon2015conflict}, while PBS improves efficiency via global partial priority orders, making it suitable for large-scale problems \citep{ma2019searching,RHCR}. The prefix "RH" added to CBS and PBS indicates that they are implemented in a rolling horizon framework for lifelong MAPF following RHCR \citep{RHCR}. PIBT performs greedy, single-step decentralized coordination with priority inheritance and backtracking to break cycles, delivering excellent scalability on large agent populations. WPPL, the winner of the 2023 League of Robot Runner Competition \citep{chan2024the}, combines PIBT \citep{PIBT} for fast initial planning with windowed MAPF-LNS \citep{li2021anytime} for refinement, balancing speed and solution quality. Their diverse trade-offs in scalability, efficiency, and solution quality ensure a comprehensive evaluation of RL-RH-PP. We exclude purely end-to-end learning-based MAPF approaches such as PRIMAL$_2$ \citep{Primal2} and Followers \citep{Follower}, as they have failed to consistently outperform non-learning-based solvers like RHCR \citep{RHCR}.

Each solver is allocated a total time budget of 1s CPU time per planning step, following \cite{jiang2024scalinglifelongmultiagentpath}. For search-based methods such as PBS and CBS, finding a feasible solution within this limited time becomes challenging, especially in high-agent-density scenarios. To address this, we apply the same repair mechanism from Section~\ref{subsec:RHPP} to any infeasible solution returned upon solver timeout. For experiments in this section, we use $K=5$ and $\beta=100$ for the Top-$K$ sampling during testing. Each reported value is averaged over 16 held-out evaluation environments with different random seeds. 

We track the total throughput of the warehouse over the $T=800$ simulation episode in \hz{16 held-out evaluation environments to monitor the training process, where each held-out environment has a different unseen random seed for initializing agent location and task assignments.} 
We focus on two primary metrics:
\begin{itemize}
    \item \textbf{Throughput per agent (TPA)}: The average number of tasks successfully completed per agent given the total $T=800$ simulation horizon. Higher throughput implies more efficient task completion under continuous goal assignments.
    \item \textbf{Total throughput}: The total number of tasks successfully completed per agent given the total $T=800$ simulation horizon, equivalent to TPA $\times$ $N$.
    \item \textbf{Solve time}: The average solve time per planning step, reflecting how quickly a solver can update paths. We track the solve time as only CPU wall time for other baselines, and CPU wall time + GPU inference time for RL-RH-PP \hz{since the RL policy is running on GPU.}
\end{itemize}

\subsection{Demonstrating the Effectiveness of RL}
\label{subsec: eft_RL}
\begin{figure*}[t]
\centering
\includegraphics[width=\textwidth]{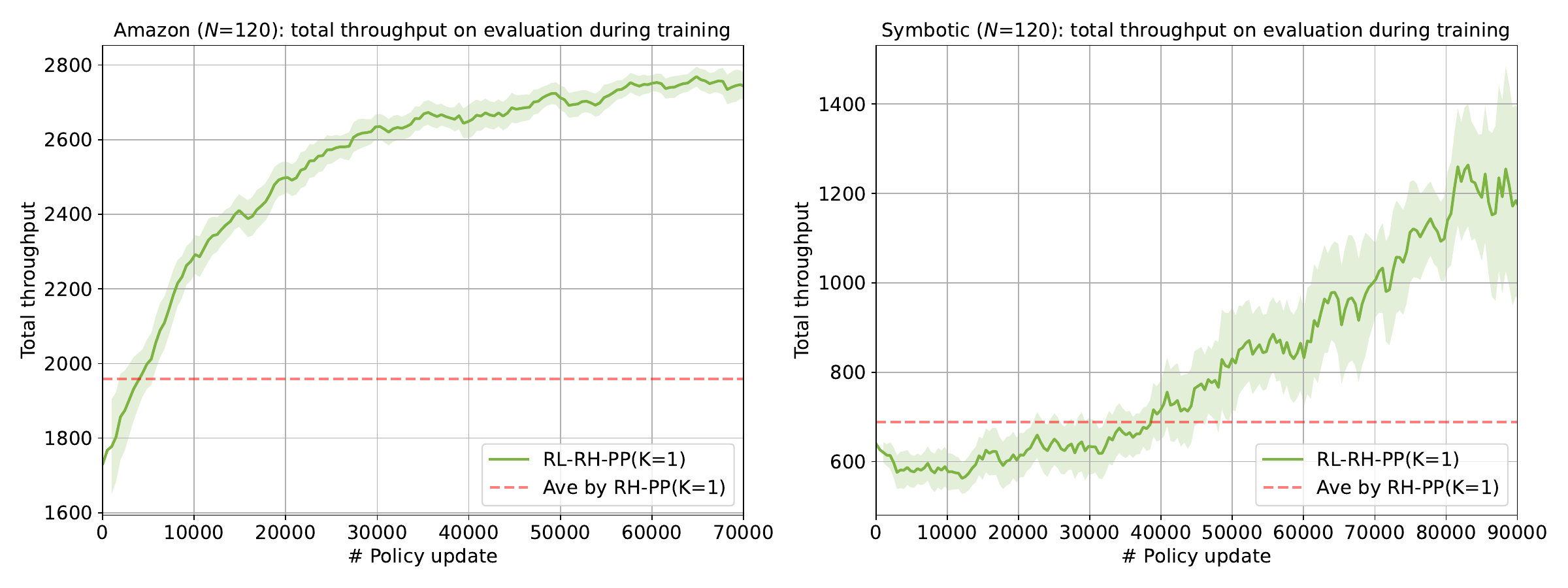}
\caption{\RABC{Ave. total throughput on held-out evaluation environments during training with $N=120$.}}
\label{fig:training}
\end{figure*}

\RABC{We first show the effectiveness of RL during training. Figure~\ref{fig:training} depicts two typical throughput curves during training, on the Amazon and Symotic maps, respectively. To better visualize the curves, we apply smoothing using a moving average of 5 consecutive policy updates, and the curves show the averaged value over the 16 evaluation environments. To make the learning progress interpretable, the training plot is augmented with a horizontal reference line representing the corresponding RH-PP baseline.}

\begin{table}[h]
\caption{\RABC{Evaluation comparison of RH-PP and RL-RH-PP with $K=1$, RL-RH-PP policy trained from scratch with $w=20$ for each setting, TPA = throughput per agent, `Ama.' for Amazon and `Sym.' for Symbotic. }}
\centering
\resizebox{0.9\textwidth}{!}{%
\begin{threeparttable}
\begin{tabular}{c@{\hspace{0.25cm}}l@{\hspace{0.1cm}}|c@{\hspace{0.15cm}}c|c@{\hspace{0.15cm}}c|c@{\hspace{0.15cm}}c}
\toprule\midrule
\multicolumn{2}{c|}{\multirow{2}{*}{{\textbf{Method}}}} & \multicolumn{2}{c|}{\textbf{$N\!=\!80$}} & \multicolumn{2}{c|}{\textbf{$N\!=\!100$}} & \multicolumn{2}{c}{\textbf{$N\!=\!120$}} \\
\multicolumn{2}{c|}{} & \textbf{TPA} $\uparrow$ & \textbf{Time(s)} $\downarrow$  & \textbf{TPA} $\uparrow$ & \textbf{Time(s)} $\downarrow$  & \textbf{TPA} $\uparrow$  & \textbf{Time(s)} $\downarrow$  \\
\midrule
\multirow{2}*{\rotatebox{90}{\textbf{Ama.}}}

& RH-PP ($K=1$) &   28.77 $\pm$ 0.91 & 0.13 & 22.49 $\pm$ 1.29& 0.19 & 16.34 $\pm$ 1.22 & 0.25 \\
& \textbf{RL-RH-PP} ($K=1$)& \textbf{30.55} $\pm$ 0.56 & 0.13 & \textbf{ 25.59} $\pm$ 1.01 & 0.20 & \textbf{22.69}  $\pm$1.07 & 0.26 \\
\midrule
\multirow{2}{*}{\rotatebox{90}{\textbf{Sym.}}} 

& RH-PP ($K=1$) & 15.86 $\pm$ 0.69  & 0.11 &  10.85 $\pm$  4.69 & 0.16 & 5.97 $\pm$ 0.95 & 0.21 \\
& \textbf{RL-RH-PP} ($K=1$)  & \textbf{17.73} $\pm$ 0.22  & 0.12 & \textbf{13.29} $\pm$ 3.44 & 0.17 & \textbf{10.74} $\pm$ 3.21 & 0.22 \\
\midrule
\bottomrule
\end{tabular}
\end{threeparttable}}
\label{tab:main_2_row_K_1}
\end{table}

\begin{table}[h]
\caption{\RABC{Evaluation comparison of RH-PP and RL-RH-PP with $K=5$, RL-RH-PP policy trained from scratch with $w=20$ for each setting, TPA = throughput per agent, `Ama.' for Amazon and `Sym.' for Symbotic. }}
\centering
\resizebox{0.9\textwidth}{!}{%
\begin{threeparttable}
\begin{tabular}{c@{\hspace{0.25cm}}l@{\hspace{0.1cm}}|c@{\hspace{0.15cm}}c|c@{\hspace{0.15cm}}c|c@{\hspace{0.15cm}}c}
\toprule\midrule
\multicolumn{2}{c|}{\multirow{2}{*}{{\textbf{Method}}}} & \multicolumn{2}{c|}{\textbf{$N\!=\!80$}} & \multicolumn{2}{c|}{\textbf{$N\!=\!100$}} & \multicolumn{2}{c}{\textbf{$N\!=\!120$}} \\
\multicolumn{2}{c|}{} & \textbf{TPA} $\uparrow$ & \textbf{Time(s)} $\downarrow$  & \textbf{TPA} $\uparrow$ & \textbf{Time(s)} $\downarrow$  & \textbf{TPA} $\uparrow$  & \textbf{Time(s)} $\downarrow$  \\
\midrule
\multirow{2}*{\rotatebox{90}{\textbf{Ama.}}}

& RH-PP ($K=5$) &   30.85 $\pm$ 0.52 & 0.65 & 25.92 $\pm$ 1.32& 0.78 & 21.74 $\pm$ 2.25 & 0.95 \\
& \textbf{RL-RH-PP} ($K=5$)& \textbf{31.80} $\pm$ 0.46 & 0.65 & \textbf{ 28.84} $\pm$ 0.35 & 0.79 & \textbf{25.56}  $\pm$0.55 & 0.96 \\
\midrule
\multirow{2}{*}{\rotatebox{90}{\textbf{Sym.}}} 

& RH-PP ($K=5$) & 16.51 $\pm$ 0.35  & 0.68 &  11.39 $\pm$  4.22 & 0.80 & 8.21 $\pm$ 2.24 & 0.96 \\
& \textbf{RL-RH-PP} ($K=5$)  & \textbf{18.38} $\pm$ 0.50  & 0.68 & \textbf{15.38} $\pm$ 1.85 & 0.82 & \textbf{11.31} $\pm$ 2.21& 0.99 \\
\midrule
\bottomrule
\end{tabular}
\end{threeparttable}}
\label{tab:main_2_row}
\end{table}

\RABC{From the plot, we observe that, initially, throughput is relatively low due to the random initialization of the policy; however, as training advances, the agent refines its prioritization strategy, potentially reducing congestion and improving path efficiency; by the later stages of training, the throughput stabilizes, suggesting that the policy has converged to an effective strategy to generate priority orders.  Compared to RH-PP, the RL-RH-PP curve rises above the RH-PP reference line early in training and remains consistently higher throughout the training process. This trend indicates that the RL policy is learning to generate more effective priority orders, leading to improved coordination among agents and more efficient task completion.}

\RABC{To further comprehensively demonstrate the consistent effectiveness of RL in enhancing RH-PP across various numbers of agents and inference samples $K$, Table~\ref{tab:main_2_row_K_1} and Table~\ref{tab:main_2_row} compares RH-PP with its learning-guided variant, RL-RH-PP, under the same sampling budget $K=1$ and $K{=}5$ with planning horizon $w{=}20$. Across all settings, RL-RH-PP achieves higher throughput per agent (TPA) at essentially the same solving time, indicating that the priority orders proposed by the RL policy are consistently better than randomly sampled orders. On Amazon map, gains are notable at $N{=}100$ and persist on $N{=}120$. On Symbotic map, improvements grow with density/scale, from a modest lift at $N{=}80$ to substantial gains at $N{=}120$, again with comparable runtime. These results (i.e., an average improvement of 25\%) demonstrate that learning to propose global priority orders significantly enhances RH-PP, especially in dense, high-traffic regimes where naive sampling struggles.}

\begin{table}[h]
\caption{\RABC{Evaluation comparison of RL-RH-PP to various baselines, RL-RH-PP policy trained from scratch with $w=20$ for each setting. TPA = throughput per agent. Best is \textbf{bolded}; second best is \underline{underlined}.} }
\centering
\resizebox{0.8\textwidth}{!}{%
\begin{threeparttable}
\begin{tabular}{c@{\hspace{0.25cm}}l@{\hspace{0.1cm}}|c@{\hspace{0.15cm}}c|c@{\hspace{0.15cm}}c|c@{\hspace{0.15cm}}c}
\toprule\midrule
\multicolumn{2}{c|}{\multirow{2}{*}{{\textbf{Method}}}} & \multicolumn{2}{c|}{\textbf{$N\!=\!80$}} & \multicolumn{2}{c|}{\textbf{$N\!=\!100$}} & \multicolumn{2}{c}{\textbf{$N\!=\!120$}} \\
\multicolumn{2}{c|}{} & \textbf{TPA} $\uparrow$ & \textbf{Time(s)} $\downarrow$  & \textbf{TPA} $\uparrow$ & \textbf{Time(s)} $\downarrow$  & \textbf{TPA} $\uparrow$  & \textbf{Time(s)} $\downarrow$  \\
\midrule
\multirow{5}*{\rotatebox{90}{\textbf{Amazon}}} 
& RH-CBS  &  4.94  $\pm$ 0.83 & 1.00  & 3.58 $\pm 0.52$ & 1.00  & 2.84 $\pm$0.29     & 1.00 \\
& RH-PBS  & 22.13 $\pm$ 7.58 & 1.00  & 4.92 $\pm$ 0.65  & 1.00  & 3.37 $\pm$ 0.26     & 1.00  \\
& PIBT  & 21.19 $\pm$ 0.58 & 0.16 & 18.62 $\pm$ 0.35 & 0.20 & 16.09 $\pm$ 0.48 & 0.27 \\
& WPPL  & \underline{29.70} $\pm$ 0.51 & 1.00 & \underline{26.60} $\pm$ 0.48 & 1.00 & \underline{23.59} $\pm$ 0.26 & 1.00 \\
& \textbf{RL-RH-PP (ours)}& \textbf{31.80} $\pm$ 0.46 & 0.65 & \textbf{28.84} $\pm$ 0.35 & 0.79 & \textbf{25.56}  $\pm$0.55 & 0.96 \\
\midrule
\multirow{5}{*}{\rotatebox{90}{\textbf{Symbotic}}} 
& RH-CBS  & 3.27 $\pm$ 0.89 & 1.00  & 2.15 $\pm$ 0.78  & 1.00  & 1.50 $\pm$ 0.45     & 1.00 \\
& RH-PBS  & \textbf{20.83} $\pm$ 1.32 & 0.75  & \underline{14.04} $\pm$ 5.39& 1.00  & 1.76  $\pm$ 1.10    & 1.00 \\
& PIBT  & 5.16 $\pm$  1.02 & 0.11 & 3.65 $\pm$ 0.77 & 0.15 &  2.67 $\pm$ 0.49 & 0.21 \\
& WPPL  & 16.29 $\pm$ 1.31 & 1.00 & 13.73 $\pm$ 1.83 & 1.00 & \underline{10.05} $\pm$ 1.33 & 1.00 \\
& \textbf{RL-RH-PP (ours)}  & \underline{18.38} $\pm$ 0.50  & 0.68 & \textbf{15.38} $\pm$ 1.85 & 0.82 & \textbf{11.31} $\pm$ 2.21& 0.99 \\
\midrule
\bottomrule
\end{tabular}
\end{threeparttable}}
\label{tab:main}
\end{table}

\subsection{Benchmarking Against Baselines}
\label{sec:benchmark}
\RABC{We now benchmark the TPA of our proposed RL-RH-PP ($K$=5) against baselines mentioned in Section~\ref{subsec:throughput & time}.
Table~\ref{tab:main} shows that RL-RH-PP achieves the best or near-best throughput per agent across both warehouse layouts and all tested scales, while maintaining inference times comparable to the strongest baselines. On the Amazon map, RL-RH-PP consistently improves over both classical solvers (RH-CBS, RH-PBS) and strong pipelines (WPPL) at similar CPU time. PIBT remains the fastest but yields clearly lower throughput. On the Symbotic map, our RL-RH-PP performs comparably to the strongest solver, RH-PBS. However, as agent density increases, the performance of RH-PBS degrades significantly, reflecting its sensitivity to congestion and scalability. In contrast, our RL-RH-PP maintains robust performance and becomes notably stronger in high-traffic regimes, surpassing all search-based methods without incurring additional inference-time cost. Overall, the proposed learning-guided RL-RH-PP achieves robust gains in coordinating robots to maximize throughput by learning to predict global priority orders, while preserving practical runtime, particularly in more complex scenarios under heavy congestion.}

\RC{A natural question is whether using step-wise decentralized coordination can match lifelong performance. Here, we address this question by focusing on the comparison of RL-RH-PP vs PIBT, which assigns and inherits priorities step-wise at each timestep without lookahead rolling-horizon planning. Empirically, RL-RH-PP outperforms PIBT across both maps and scales in Table~\ref{tab:main}, with the gap widening at higher densities on the Symbotic layout. We attribute this to RL-RH-PP’s rolling-horizon planning and learned global priority orders, which explicitly account for cascading, long-horizon interactions. In contrast, PIBT’s per-step, myopic updates lack lookahead and are more sensitive to congestion, leading to reduced throughput despite its lower runtime.}

\subsection{Generalizing to Different Scenarios}
\label{subsec:gen_anay}
\begin{figure*}[t]
\centering
\includegraphics[width=\textwidth]{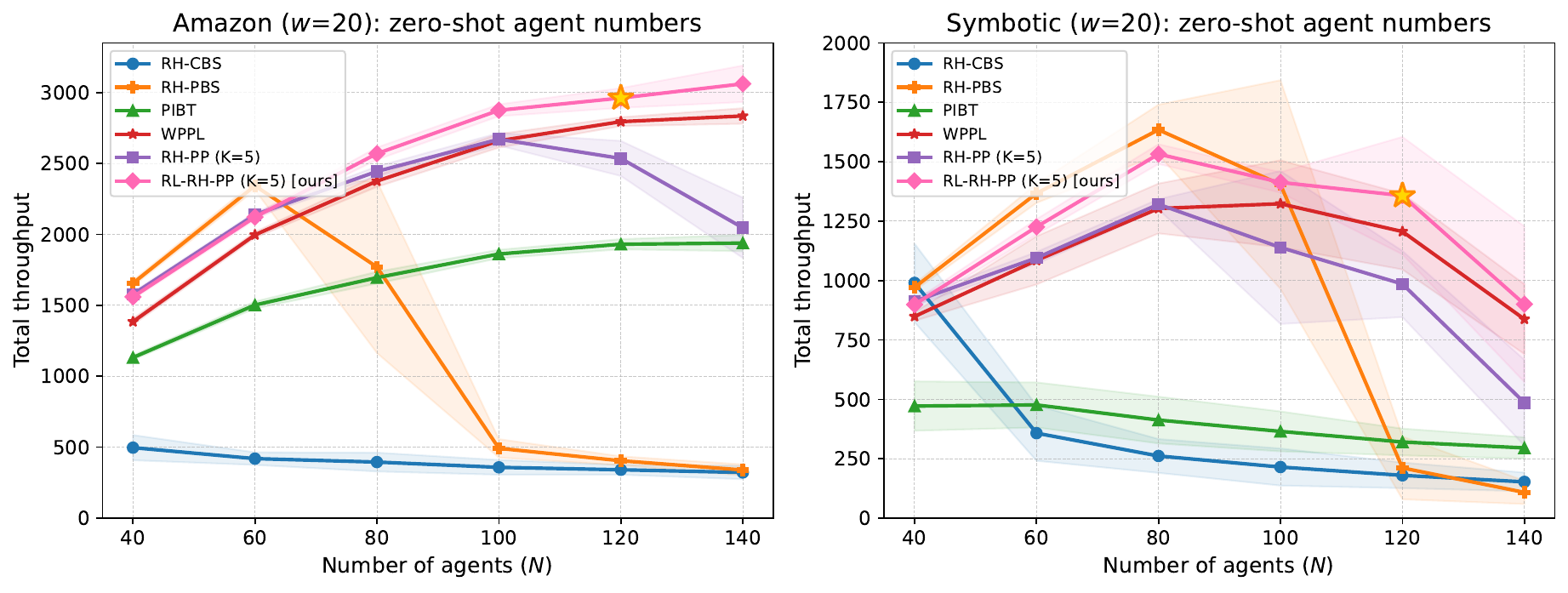}
\caption{\RABC{Total throughput versus agent number $N$ in zero-shot transfer evaluation at $w=20$. The base model (yellow star) is trained at $N=120$ and $w=20$, and evaluated zero-shot at other $N$.}}
\label{fig:zeroshot_agents}
\end{figure*}
\RALL{Since the proposed method involves learning, its zero-shot generalization performance is evaluated across diverse setups, including variations in the number of agents, planning window size, and unseen map layouts.}

\subsubsection{Generalization across different numbers of agents}
\label{subsec:gen_agent}
\RABC{
We assess zero-shot transfer across agent populations using a single base policy trained at $N=120$ and planning horizon $w=20$. At inference time, we fix the same planning horizon and vary $N$ without any retraining, keeping all planner settings identical across methods for fairness. Figure~\ref{fig:zeroshot_agents} plots total throughput as a function of $N$.}

\RABC{
We report the results in Figure~\ref{fig:zeroshot_agents}, where similar performance can be observed compared to the in-distribution performance shown in Section~\ref{sec:benchmark}, indicating consistent robust performance of our proposed RL-RH-PP.
Specifically, on the Amazon maps, the strong search-based baseline RH-PBS slightly outperforms the proposed method at lower agent densities. However, as the number of agents increases, its effectiveness drops significantly, while our RL-RH-PP maintains the best performance throughout. RH-PP remains a strong non-learning reference with random orders; however, our RL-RH-PP consistently improves upon it even with zero-shot generalization, indicating that the policy has learned generalizable knowledge on predicting higher-quality global priority orders, especially under heavier congestion. Moreover, it is notable that as the number of agents increases, the performance of the original RH-PP declines, while RL-RH-PP continues to improve. This suggests that RL-RH-PP is less constrained by the performance ceiling of the RH-PP backbone. The RL component potentially enables better adaptation under higher agent densities, partially mitigating limitations of the underlying RH-PP. This is a compelling case for learning-guided optimization. On the more constrained Symbotic maps, throughput is lower and decreases as the number of agents increases for all methods, likely due to inherent environmental patterns. Compared with RH-PP, our proposed method consistently achieves better performance despite operating under zero-shot generalization. The value of robust prioritization becomes increasingly evident with larger agent populations, suggesting that the learned policy captures congestion patterns that scale with crowding. WPPL serves as a strong baseline on both maps, and RL-RH-PP outperforms it under almost all settings. Overall, a policy trained only at $N=120$ generalizes smoothly to both smaller and larger populations without any retraining.}

\subsubsection{Generalization across different planning windows}
\label{subsec:gen_horizon}
\begin{figure*}[t]
\centering
\includegraphics[width=\textwidth]{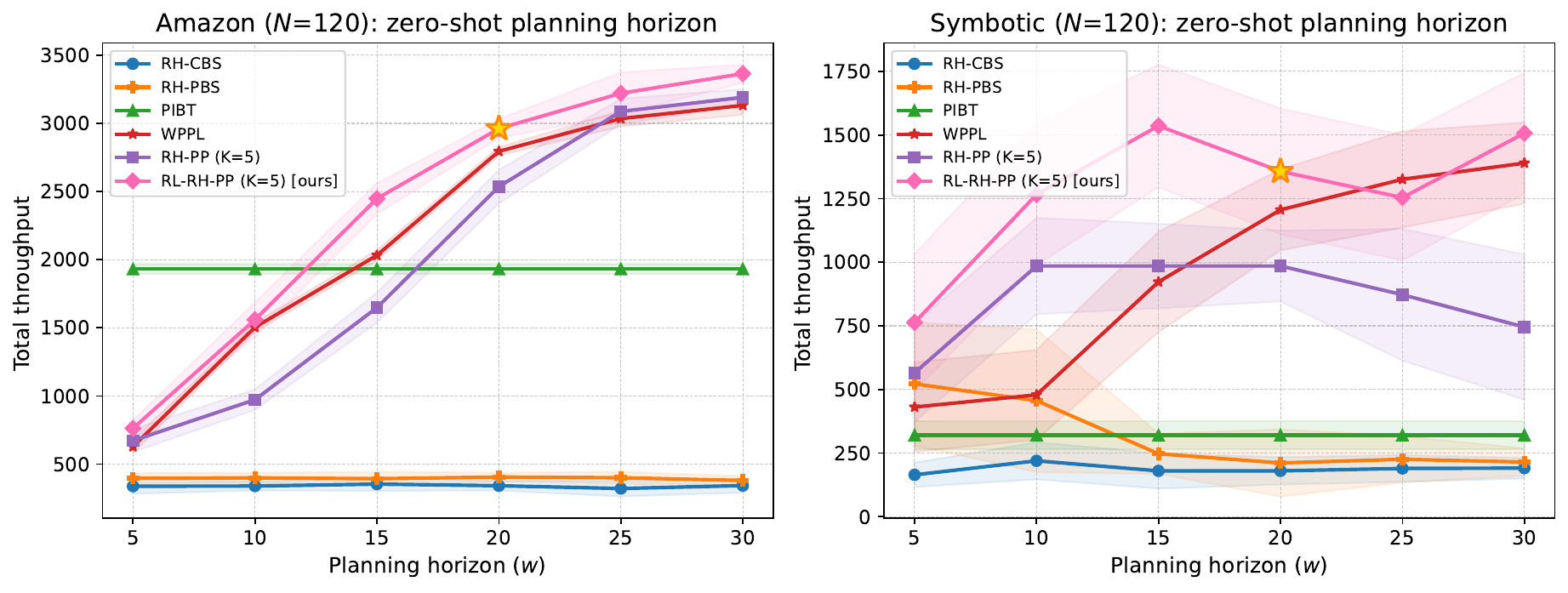}
\caption{\RABC{Total throughput versus planning horizon $w$ in zero-shot transfer evaluation at $N=120$. The base model (yellow star) is trained at $N=120$ and $w=20$, and evaluated zero-shot at other $w$.}}
\label{fig:zeroshot_horizon}
\end{figure*}

\RABC{We next examine zero-shot transfer across planning horizons by fixing $N=120$ and varying $w$ at inference time, using the same base policy trained at $w=20$. For comparability, RH-CBS, RH-PBS, RH-PP, and WPPL use the same window in each experiment. Figure~\ref{fig:zeroshot_horizon} shows total throughput versus $w$.} 

\RABC{
As can be seen from the results, on Amazon maps, increasing $w$ generally benefits all methods by providing more lookahead information. RL-RH-PP is competitive at small windows and remains at or near the top as the window grows, leveraging the longer preview to coordinate agents more proactively. RH-PP also improves with larger $w$, but the learned order generator usually yields additional gains over random orders, underscoring the advantage of learned global prioritization even when the planner sees further ahead. On Symbotic, longer windows are more consequential: RL-RH-PP capitalizes on larger $w$ to anticipate corridor conflicts and relieve pressure in cross-aisles earlier within the horizon, surpassing baselines in the higher-$w$ regime. These results indicate that a policy trained once at $w=20$ adapts well to both smaller and larger windows at test time. Again, it is observed that as the planning horizon increases, RL-RH-PP is able to reverse the declining trend of RH-PP, providing further evidence of its ability to bypass inherent limitations of RH-PP for robust learning-guided optimization. We further evaluate mixed zero-shot transfer where both the agent population $N$ and the planning window $w$ vary simultaneously; the results are reported in the appendix~\ref{apd: mixed_agent_horizon}}.

\subsubsection{Remark on best configuration for each method}

As observed in our experiments, for a given fixed map, each solver exhibits its peak performance in terms of throughput per bot at a specific value of $N$ and $w$. The total throughput of the warehouse is computed as the throughput per bot multiplied by $N$.  From a practical perspective of an autonomous warehouse service provider, it is essential to not only optimize throughput per agent but also to identify the ideal configuration of agent density and planning horizon to maximize overall operational throughput. \RA{To maximize the total throughput of the warehouse, we first identify the highest potential total throughput achievable by each solver under different configurations $(N,w)$ from experiments in Section~\ref{subsec:gen_agent} and Section~\ref{subsec:gen_horizon}, and then compare their performance against this. The results are summarized in Table~\ref{tab:total throughput}.}

\begin{table}[h]
\caption{\RA{Highest total throughput of each methods under the corresponding configurations $(N,w)$  (TP = throughput).}}
\centering
\small
\resizebox{0.4\textwidth}{!}{%
\begin{threeparttable}
\begin{tabular}{c@{\hspace{0.25cm}}l@{\hspace{0.1cm}}|c|c}
\toprule\midrule
\multicolumn{2}{c|}{\textbf{Method}} & \textbf{Total TP} $\uparrow$ & \textbf{Optimal $(N,w)$} \\
\midrule
\multirow{5}*{\rotatebox{90}{\textbf{Amazon}}} 
& RH-CBS  & 512 $\pm$ 83 & (40,10) \\
& RH-PBS  & 2343 $\pm$ 30 & (60,20) \\
& PIBT  & 1939 $\pm$ 53 & (140,1) \\ 
& WPPL  & 3132 $\pm$ 68 & (120,30) \\
& RH-PP  & 3170 $\pm$ 60& (120,30) \\
& \textbf{RL-RH-PP} & \textbf{3364} $\pm$ 67  & (120,30)\\
\midrule
\multirow{5}{*}{\rotatebox{90}{\textbf{Symbotic}}} 
& RH-CBS  & 990  $\pm$ 165 & (40,20) \\
& RH-PBS  & 1666  $\pm$ 105 & (80,20) \\
& PIBT  & 477 $\pm$ 97 & (60,1) \\ 
& WPPL  & 1389  $\pm$ 159 & (120,30) \\
& RH-PP  & 1611  $\pm$ 57 & (100,10) \\
& \textbf{RL-RH-PP} & \textbf{1740}  $\pm$ 32 & (100,10) \\
\midrule
\bottomrule
\end{tabular}
\end{threeparttable}}
\label{tab:total throughput}
\end{table}

Our proposal RL-RH-PP achieves the highest total throughput in both warehouse maps, demonstrating its robustness and its ability to generalize effectively across different scenarios. By identifying the optimal configuration for each solver, our analysis provides practical insights into maximizing total throughput.

\subsubsection{Generalization across different maps}
\label{subsec:gen_maps}
\begin{figure*}[h]
\centering
\includegraphics[width=0.95\textwidth]{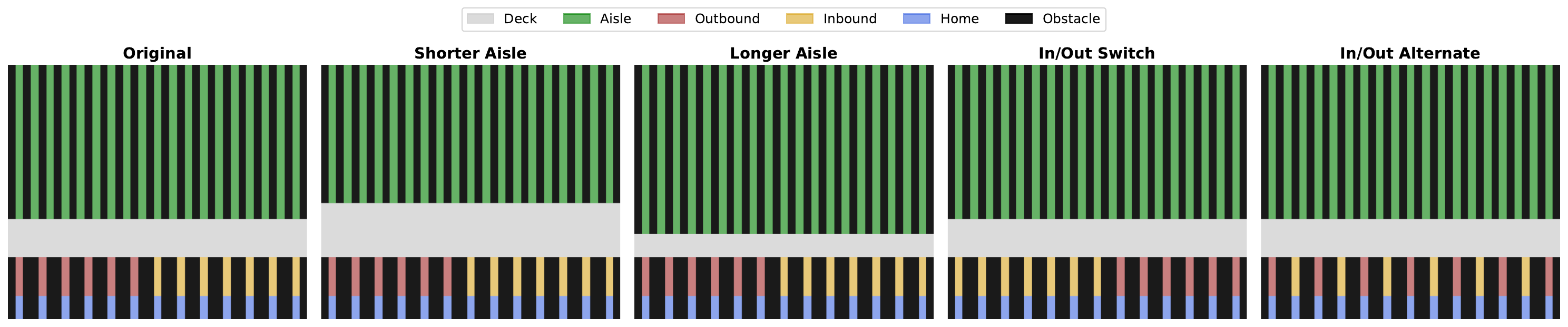}
\caption{\RB{Symbotic-style layout family: Original (training layout), Shorter Aisle (uniformly shortened aisles), Longer Aisle (uniformly lengthened aisles), In/Out Switch (inbound and outbound docks swapped), and In/Out Alternate (inbound/outbound docks interleaved along the perimeter). All share the same map size $(W,H)$.}}
\label{fig:symbotic_variants}
\end{figure*}

\begin{figure*}[h]
\centering
\includegraphics[width=0.8\textwidth]{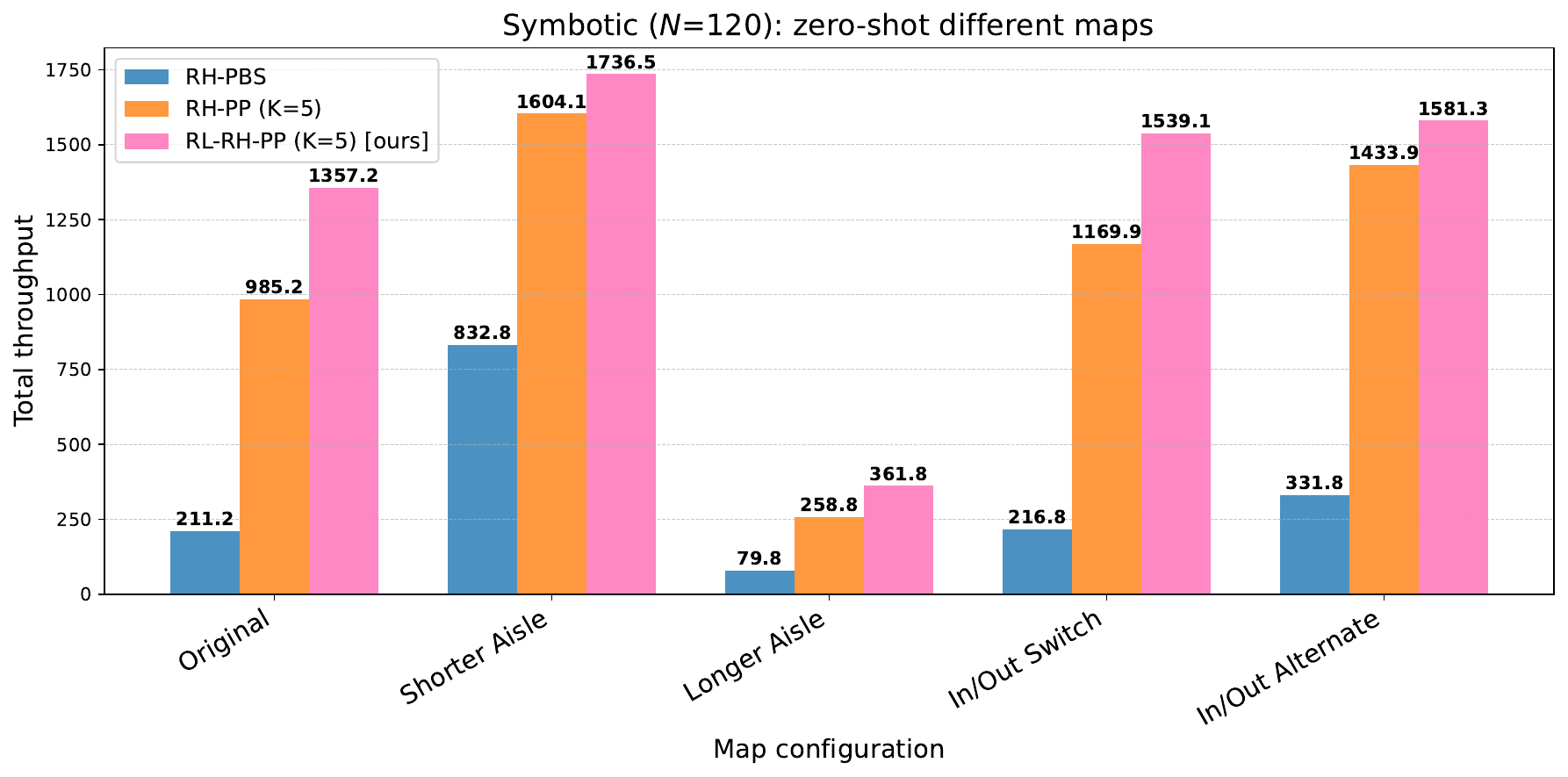}
\caption{\RB{Total throughput versus different Symbotic maps in zero-shot transfer evaluation at $N=120$ and $w=20$. The base model is trained at $N=120$ and $w=20$ on the Original map.}}
\label{fig:gen_total_throughput}
\end{figure*}

\RB{It is also of interest to evaluate the extent to which the the trained policy generalizes across unseen map layouts. To probe cross-layout robustness, we derive four Symbotic-style variants from the base warehouse while preserving the same map size $(W,H)$ so that absolute indices remain valid for our encoder. Those map variants are shown in Fig.~\ref{fig:symbotic_variants}. We choose Symbotic rather than Amazon because its higher obstacle density, narrower aisles, and more frequent intersections create stronger congestion externalities, yielding a stricter test for cross-layout robustness. Specifically, we include the following maps:
\begin{itemize}
  \item Original: the unmodified training layout.
  \item Shorter Aisle: aisles uniformly shortened, increasing cross-aisle contention near intersections.
  \item Longer Aisle: aisles uniformly lengthened, creating longer corridors and queueing at choke points.
  \item In/Out Switch: inbound and outbound docks swapped to reverse macroscopic flow.
  \item In/Out Alternate: inbound/outbound docks interleaved to induce persistent opposing perimeter streams.
\end{itemize}
We choose a single base policy trained on the Original layout at $N=120$ and $w=20$ with fixed inference configuration, then evaluate it zero-shot on each variant. Figure~\ref{fig:gen_total_throughput} reports total throughput at $N=120$ and $w=20$ across the original and four variant maps. The zero-shot policy consistently exceeds RH-PP and significantly surpasses RH-PBS on every variant, with the largest gains on layouts that intensify congestion externalities (Longer Aisle, In/Out Alternate) and smaller but positive margins on layouts closer to the training geometry (Shorter Aisle, In/Out Switch). These results suggest that learning-guided prioritization provides useful cross-layout robustness within same-footprint reconfigurations, while full map-agnostic generalization (e.g., different map size) is left for future work. We also evaluate a mixed zero-shot setting where the agent population $N$ varies across the different layout variants without retraining; results are provided in the appendix~\ref{apd: mixed_agent_horizon}.}

\subsubsection{Remark on why zero-shot transfer works}
\label{subsec:why_zeroshot}
\RB{These experiments indicate that the policy, trained under a fixed configuration, generalizes effectively within the same map to different agent densities and planning horizons. A key reason is the position-embedding design: each drivable location has a trainable embedding, and the policy learns to interpret spatial and temporal relationships by attending over sequences of these embeddings along agents’ paths. Because the map remains unchanged, the same embedding dictionary is reused at test time; varying $N$ or $w$ simply changes which sequences are presented, not the underlying representation. Coupled with attention layers that merge spatial and temporal information, this yields robustness to workload and horizon changes. Cross-map zero-shot transfer is harder with absolute embeddings and is addressed in our layout-variant study above; fully map-agnostic encoders are an avenue for future work.}

\subsection{Anytime Behavior of RL-RH-PP}
\label{subsec: anytime}
\begin{figure*}[t]
\centering
\includegraphics[width=0.85\textwidth]{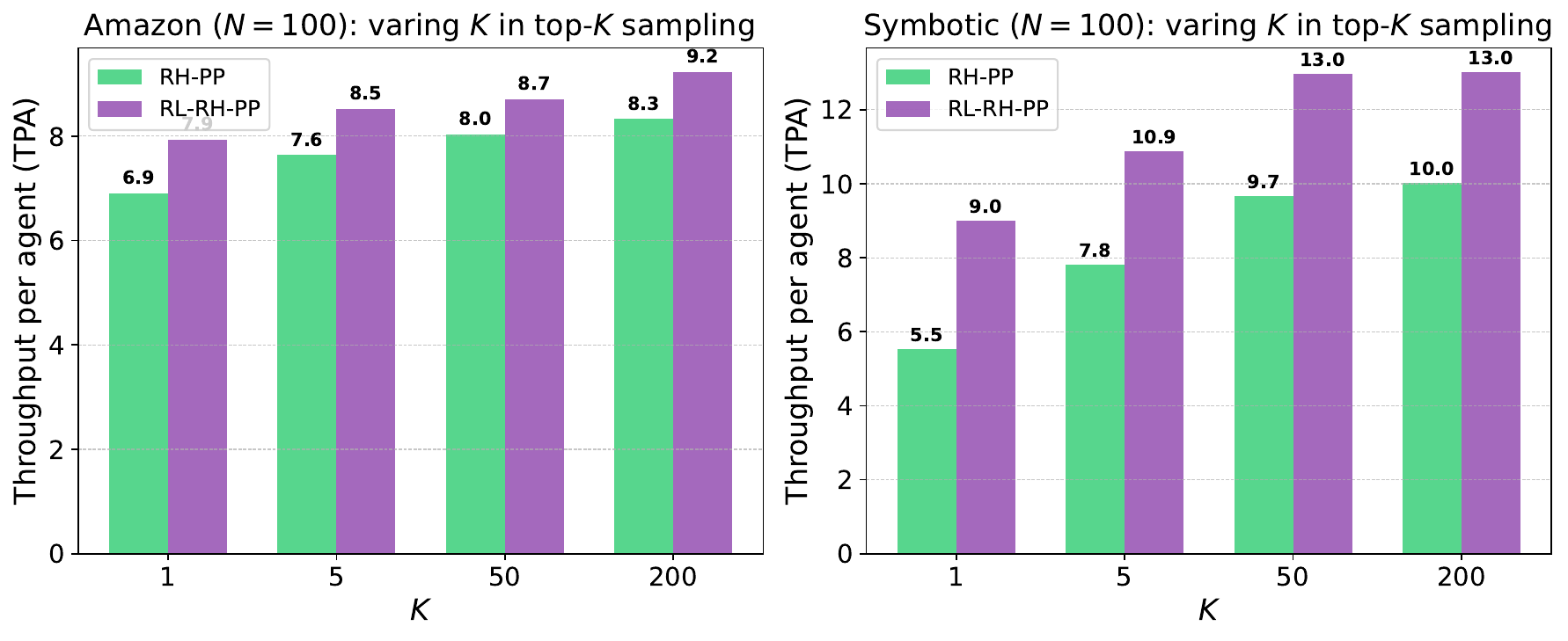}
\caption{\RABC{Throughput per agent (TPA) at evaluation vs $K$, evaluated with $N=100$}}
\label{fig:TBA_K_amazon_symbotic_100}
\end{figure*}

\RABC{With Top-$K$ sampling, we can increase the number of sampled priority orders $K$ to evaluate more candidates, making both RH-PP and RL-RH-PP anytime solvers. We first study their anytime behavior by examining how the final selected order depends on $K$. Specifically, we evaluate $K \in \{1, 5, 50, 200\}$ with a planning horizon $w=5$ on both maps. We fix $w=5$ to avoid excessive runtime at large $K$ (e.g., $K=200$). Figures~\ref{fig:TBA_K_amazon_symbotic_100} illustrate the relationship between $K$ and throughput per agent (TPA) at $N=100$, and we save the result for $N=80$ in appendix~\ref{apd: beta}.}

\label{subsec:sampling_K}

\RABC{When only a single priority order is sampled ($K=1$), RL-RH-PP consistently outperforms RH-PP across all settings. This is expected, as the RL policy is trained by sampling a single priority order ($K=1$), thereby specializing in generating high-quality solutions using that order. As $K$ increases, by leveraging learned autoregressive decoding strategy, RL-RH-PP effectively shrinks the search space for finding optimal priority orders,enabling more efficient planning compared to naive, uninformed sampling. We also observe an improvement in throughput per agent for both methods as $K$ increases, demonstrating that performance can be further enhanced by exploring a broader set of priority orders.}

\begin{figure*}[t]
\centering
\includegraphics[width=0.9\textwidth]{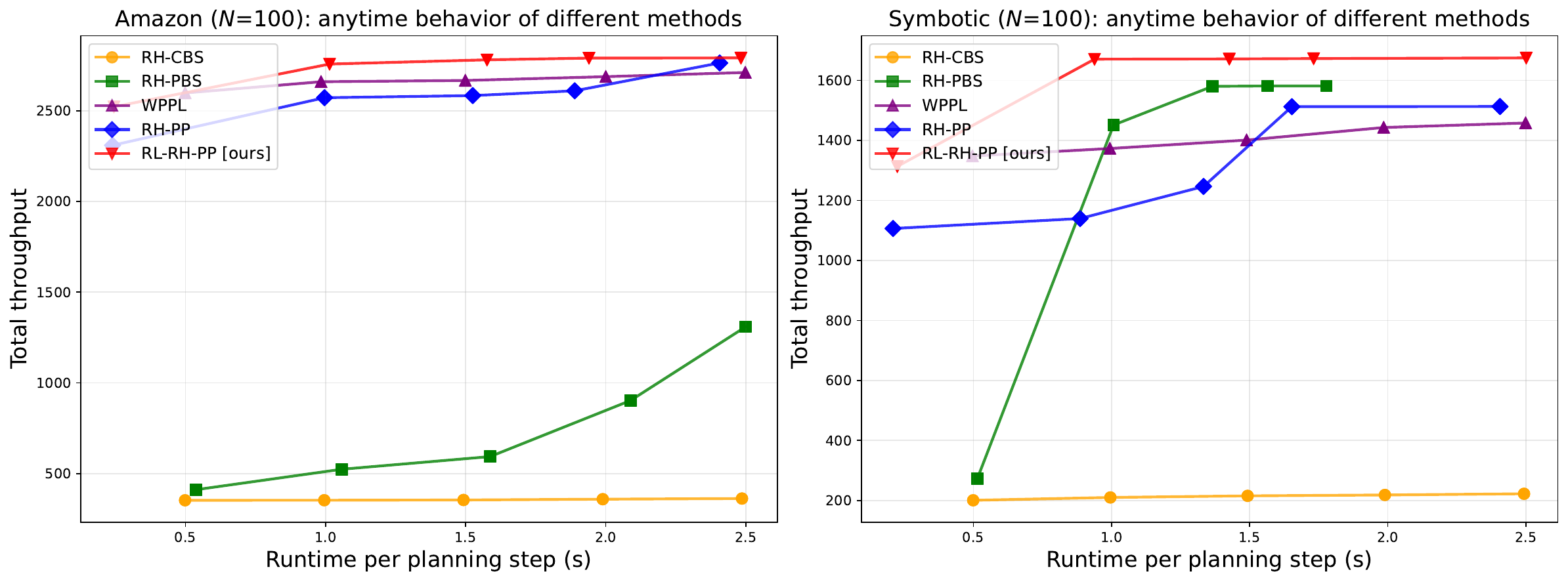}
\caption{\RA{Total throughput vs runtime budget of all methods, evaluated with $N=100$ and $w=20$}}
\label{fig:pareto_frontier}
\end{figure*}

\RA{We next quantify the runtime--quality trade-off across methods on each map with. For fairness, we sweep one knob per method that primarily controls compute: the number of sampled orders $K$ for RH-PP/RL-RH-PP, the LNS time budget for WPPL, and the search-time budget for RH-CBS/RH-PBS. For every configuration, we sweep the computational budget from $\{$0.5s, 1.0s, 1.5s, 2.0s, 2.5s$\}$ CPU time, and we report average total throughput and average end-to-end runtime per planning step on identical hardware. Fig.~\ref{fig:pareto_frontier} show that increasing $K$ yields predictable anytime behavior for RL-RH-PP. Moreover, RL-RH-PP contributes competitive points to the frontier alongside baselines, illustrating that learning to propose priority orders provides favorable quality at practical runtimes.}

\RABC{Our current implementation of Top-$K$ evaluation is executed serially on a CPU, leading to a linear increase in computation time with $K$. Table~\ref{tab:runtime_breakdown} reports the average time to (i) generate a priority order with the trained policy in inference mode and (ii) evaluate that order by running prioritized planning within RL-RH-PP. We have moved order generation to a batched, GPU-parallel implementation, so for large $K$ the dominant cost is the CPU-bound Top-$K$ evaluation. For example, at $K=200$, generating all orders takes just over 32 ms, whereas evaluating them takes about 9 s. Future work could optimize solving efficiency by leveraging multi-threading techniques to parallelize Top-$K$ evaluation, significantly reducing computation overhead and improving real-time applicability.}

\begin{table}[t]
\centering
\caption{\RA{Per priority order runtime breakdown, evaluated at Symbotic map with $N=100$} }
\label{tab:runtime_breakdown}
\begin{tabular}{lc}
\toprule
Component & Time (ms) \\
\midrule
Priority order generation (network forward pass) & 32 \\
Priority order evaluation (run prioritized planning)     & 46 \\
\bottomrule
\end{tabular}
\end{table}

\subsection{Ablation Study}
\label{sec:ablation} \RALL{In this section, we assess the effectiveness of our proposal through a series of ablation studies.}

\subsubsection{Effects of the feasibility handling schemes of the priority order}
\RABC{Prioritized planning is not complete, hence a selected priority order can occasionally yield an infeasible initial plan. As described in Section~\ref{sec:Method}, two approaches are adopted. First, within the RH-PP backbone, a heuristic scoring function guides selection among the Top-$K$ sampled orders. Second, penalties are incorporated into the reward design of RL-RH-PP training to encourage the generation of feasible orders. We now ablate the effects of both the above designs.}

\RABC{
We first study the effect of the infeasibility heuristic measure in the Top-$K$ selector. Recall that in Eq.~\ref{eq: cost}, RH-PP evaluates each sampled total order \(\prec_k\) by 
$
\mathrm{cost}(\prec_k)=\sum_{i=1}^N \big(e_i+\beta\, s_i\big)
$,
where \(e_i\) is the initial path length under \(\prec_k\), \(s_i\in\{0,1\}\) indicates whether agent \(i\) was forced onto its shortest path (making the initial solution infeasible), and \(\beta\) penalizes such infeasibility. We study how the infeasibility penalty \(\beta\) affects the quality of the final selected priority order. We sweep \(\beta\in\{0,1, 10,100\}\) for both RH-PP and RL-RH-PP. Because the order selection in Eq.~\ref{eq: cost} is applied only at inference, we do not retrain the RL policy for each \(\beta\). We set $K=10$ to include more priority orders under consideration to create more challenging short preview horizon life-long planning, while holding all other settings fixed as the experiments in Table~\ref{tab:main_2_row}.  We report throughput per agent (TPA) under $N=100$ in both maps in the Figures.~\ref{fig:TBA_beta_amazon_symbotic}, while saving the result for $N=80$ in appendix~\ref{apd: beta}. We observe a clear upward trend in TPA as \(\beta\) increases for both RH-PP and RL-RH-PP. A larger \(\beta\) raises the cost of candidate orders in which many agents are forced onto their unplanned shortest paths, so the selector prefers orders with fewer such agents. This choice reduces downstream repair, early conflicts, and wait-induced detours, yielding higher TPA.}

\begin{figure*}[h]
\centering
\includegraphics[width=0.85\textwidth]{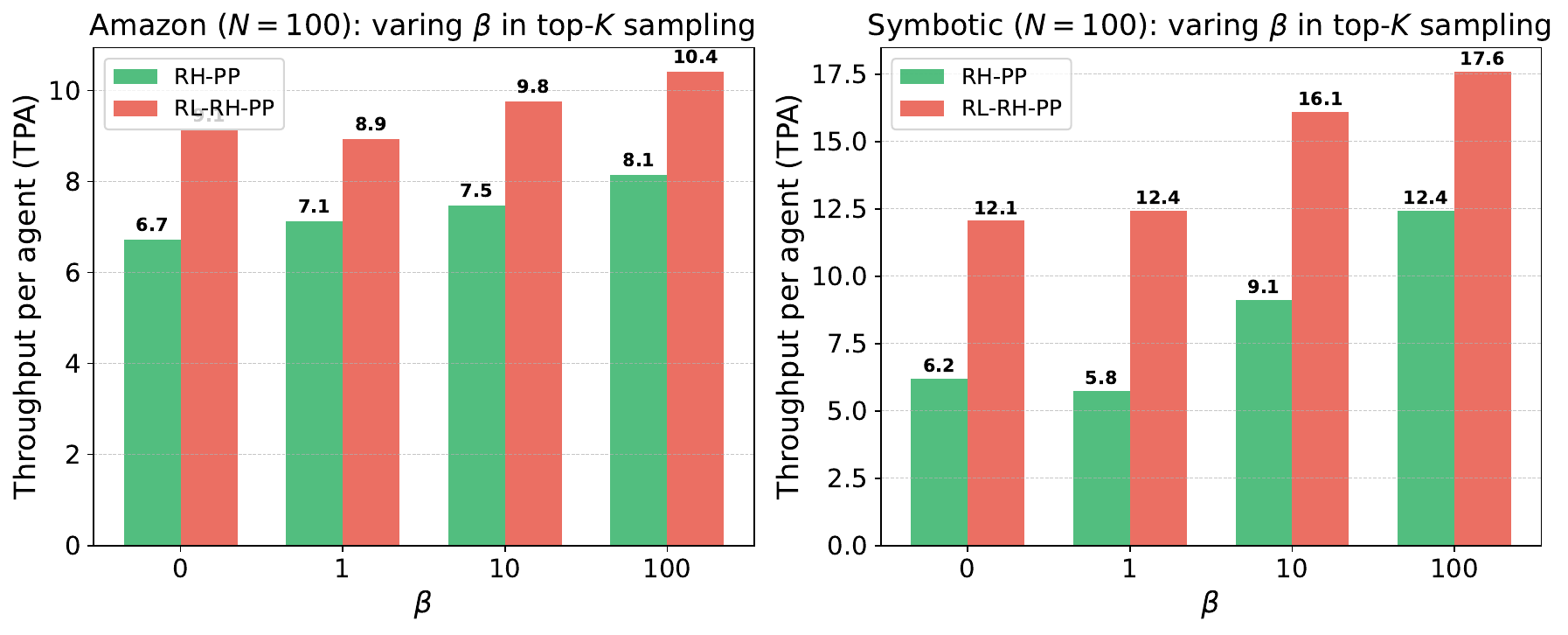}
\caption{\RABC{Throughput per agent (TPA) at evaluation vs $\beta$, evaluated with $N=100$}}
\label{fig:TBA_beta_amazon_symbotic}
\end{figure*}

\RABC{
Next, we study whether our penalty in the reward term truly boosts the feasibility of the learned policy. We quantify this by the percentage of planning steps whose selected order is infeasible (lower is better), and report the values in Table~\ref{tab:infeasibility}. For comparability, we report results from the same experiments in Table~\ref{tab:main_2_row}. Across maps and agent densities, RL-RH-PP yields a lower infeasibility rate than RH-PP, consistent with the reward term that penalizes infeasibility (the $\sigma s_{i,t}$ in Eq.~\ref{eq:rl_reward}).}

\begin{table}[h]
\caption{\RABC{Percentage of steps where the selected order yields an infeasible initial plan (lower is better), tracked from same experiments in Table~\ref{tab:main_2_row}, `Ama.' for Amazon and `Sym.' for Symbotic.}}
\centering
\resizebox{0.55\textwidth}{!}{%
\begin{threeparttable}
\begin{tabular}{c@{\hspace{0.25cm}}l|c|c|c}
\toprule\midrule
\multicolumn{2}{c|}{\textbf{Method}} & \textbf{$N\!=\!80$} & \textbf{$N\!=\!100$} & \textbf{$N\!=\!120$} \\
\midrule
\multirow{2}*{\rotatebox{90}{\textbf{Ama.}}} 
& RH-PP ($K=5$) &   8.1\% $\pm$ 2.1\% & 39\% $\pm$ 5.1\% & 88\% $\pm$ 4.3\% \\
& \textbf{RL-RH-PP} ($K=5$) & \textbf{7.8\%} $\pm$ 1.9\% & \textbf{28\%} $\pm$ 5.0\% & \textbf{60\%}  $\pm$ 6.2\% \\
\midrule
\multirow{2}{*}{\rotatebox{90}{\textbf{Sym.}}} 
& RH-PP ($K=5$) & 83\% $\pm$ 3.9\% &  97\% $\pm$  1.1\% & 99\% $\pm$ 0.1\% \\
& \textbf{RL-RH-PP} ($K=5$) & \textbf{34\%} $\pm$ 4.8\% & \textbf{84\%} $\pm$ 3.5\% & \textbf{97\%} $\pm$ 0.3\% \\
\midrule
\bottomrule
\end{tabular}
\end{threeparttable}}
\label{tab:infeasibility}
\end{table}

\begin{figure*}[t]
\centering
\includegraphics[width=1\textwidth]{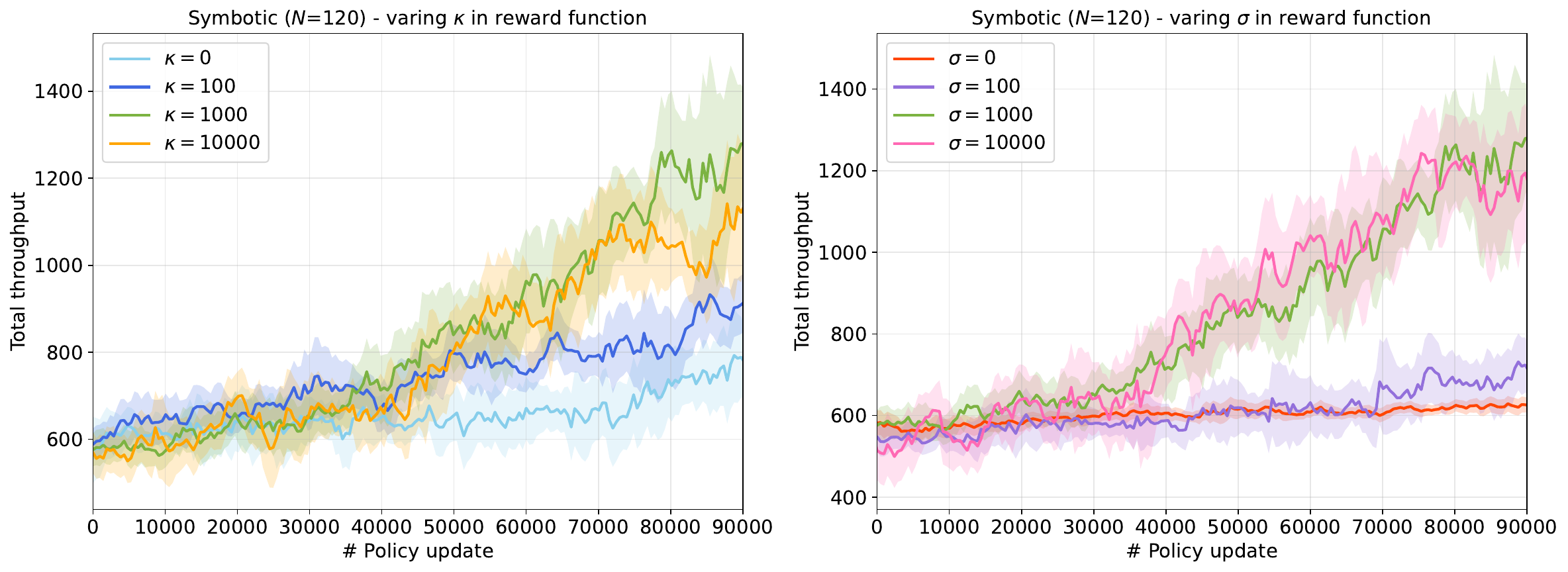}
\caption{\RB{Total throughput during training under different congestion ($\kappa$) and infeasibility ($\sigma$) weights in reward function Eq.~\ref{eq:rl_reward}, evaluated on-the-fly with $N=120$}}
\label{fig:ablation_kappa_sigmal}
\end{figure*}

\RB{\subsubsection{Effect of the penalty weights in the reward function}
\label{subsubsec:pentaly weights}
We have weighted reward terms in our reward function (Eq.~\ref{eq:rl_reward}), and now we study the effect of these penalty weights on the training process.  Recall that the designed reward function (Eq.~\ref{eq:rl_reward}) used for RL training is as follows: $
R(\boldsymbol{o}_t,\boldsymbol{a}_t)
= - \frac{1}{N} \Big( \sum_{i=1}^{N} d_{i,t} \;+\; \kappa\, c_{i,t} \;+\; \sigma\, s_{i,t} \Big)$,
where $d_{i,t}$ is the average Manhattan distance from agent $i$ to its queued goals, $c_{i,t}\in\{0,1\}$ indicates that agent $i$ has all-wait actions for the next $h$ steps (i.e., congested), and $s_{i,t}\in\{0,1\}$ indicates that no feasible path can be found for agent $i$ under the current priority. $\kappa$ and $\sigma$ are weighting factors. We study how the congestion weight $\kappa$ and the infeasibility weight $\sigma$ affect the total throughput of the trained RL-RH-PP policy. Because these weighting factors shape the training objective, we re-train RL-RH-PP from scratch for $\kappa\in\{0,100,1000,10000\}$ while fixing $\sigma=1000$, and for $\sigma\in\{0,100,1000,10000\}$ while fixing $\kappa=1000$, under identical training and evaluation settings as the experiments for $N=120$ in Table~\ref{tab:main_2_row}. Our primary metric is total throughput. To stress-test decongestion behavior and make the effect of $\kappa$ and $\sigma$ most visible, we run this ablation on the Symbotic map, which induces sustained congestion and frequent aisle conflicts. As shown in Fig.~\ref{fig:ablation_kappa_sigmal}, increasing either weight generally improves both learning speed and the final throughput plateau up to a broad optimum around $1000$. In the congestion sweep, $\kappa=0$ underperforms, $\kappa=100$ improves stability, and $\kappa=1000$ attains the best overall curve, while $\kappa=10000$ yields no further gains. In the infeasibility sweep, the trend is more pronounced: performance is more sensitive to $\sigma$, with $\sigma=0$ lagging, $\sigma=100$ narrowing the gap, $\sigma=1000$ achieving the highest plateau, and $\sigma=10000$ offering little additional benefit. These results indicate that both penalties are useful for throughput in dense regimes, that the infeasibility weight is the more sensitive knob, and that excessively large values are unnecessary. We therefore set $\kappa=1000$ and $\sigma=1000$ in our main experiments.}

\subsubsection{Effects of the temporal and spatial attention}
\begin{figure*}[t]
\centering
\includegraphics[width=\textwidth]{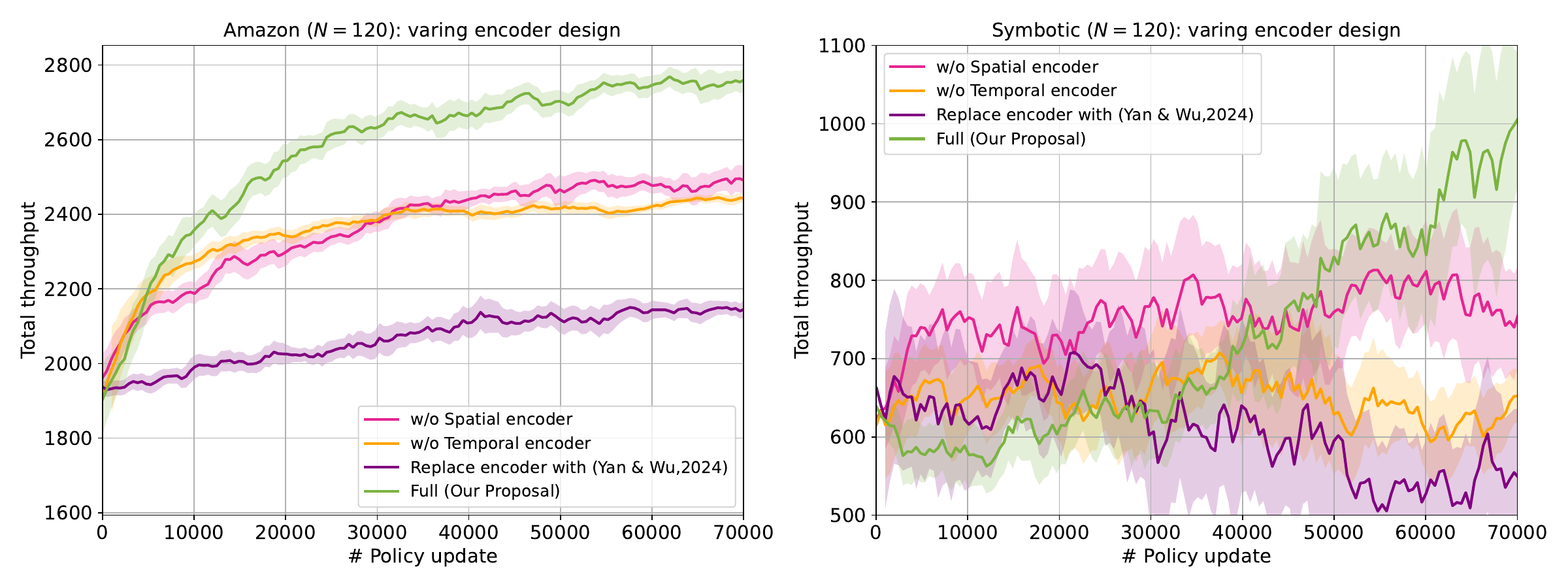}
\caption{\RB{Total throughput during training for encoder ablation study, evaluated on-the-fly with $N=120$}}
\label{fig:ablation_encoder}
\end{figure*}

\RB{To isolate the contribution of the encoder’s temporal and spatial attention design, we perform the following ablation experiments by training four encoder variants from scratch under identical budgets, seeds, and inference settings (decoder unchanged): (i) Full (ours), (ii) w/o Temporal encoder (temporal attention replaced by MLPs), (iii) w/o Spatial encoder (spatial attention replaced by MLPs), and (iv) Replace the entire encoder with the architecture of Yan \& Wu \citep{yan2024neural} (intra-path attention + 3D conv). We report total throughput during training on Amazon and Symbotic at $N{=}120$, as shown in Fig.~\ref{fig:ablation_encoder}.}

\RB{Across both maps, removing either attention block lowers learning speed and the final plateau relative to the Full model, confirming that both temporal and spatial attention are beneficial. On Amazon, the differences are modest: dropping temporal attention slows down early learning and ends slightly below Full, while dropping spatial attention yields a similarly small decrease—consistent with the map’s wider corridors and fewer bottlenecks. On Symbotic, however, the contrast is sharper. Without temporal attention, performance degrades substantially, highlighting the need to capture long-range interactions to resolve aisle congestion. Removing spatial attention also hurts, though less dramatically, suggesting that both components matter but temporal attention is particularly critical in constrained, high-density environments.}

\RB{Replacing the entire encoder with Yan \& Wu’s  \citep{yan2024neural} yields consistently lower throughput on Amazon and fails to learn a competitive policy on Symbotic. We attribute this drop to the limitations of 3D convolution, which emphasizes local interactions within a fixed receptive field: it captures short-range dependencies but struggles with global agent–agent interactions needed for long-term coordination. In contrast, our spatial attention computes agent-wise weights over the entire frame, enabling global context and more informed decisions, which improves coordination and overall throughput in lifelong MAPF.}

\subsubsection{Compare with other priority order assignment heuristics}
\RABC{To demonstrate the effectiveness of our learning-guided heuristic for priority ordering, we compare the priority orders sampled from Top-$K$ sampling in RL-RH-PP against those constructed using rule-based methods. Specifically, we evaluate the distance-query heuristic used in \cite{Berg2005,ma2019searching}, where an agent's priority is determined by the length of its shortest path from its current location to its goal. The intuition behind this heuristic is to minimize the maximum arrival time by prioritizing agents with longer paths, allowing them to move unhindered, while those with shorter paths can afford delays to yield to higher-priority agents. To integrate this heuristic into RH-PP, we replace the Top-$K$ sampling mechanism with a direct priority assignment based on the distance heuristic, constructing a total priority order at each planning step. We refer to this ablation variant as Distance Query Rolling Horizon Prioritized Planning (DQ-RH-PP).}

\FN{We do not include \cite{ZhangSoCS22} as a baseline because, in the one-shot setting it targets, its learned ordering underperforms the distance-query heuristic  \citep{Berg2005,ma2019searching} reported in the same literature. To provide a stronger and directly applicable comparator under our rolling-horizon protocol, we evaluate the DQ-based variant, DQ-RH-PP (Table~\ref{tab:heuristic_compare})}.

\begin{table}[h]
\caption{\RABC{Compare with other priority order assignment heuristics ($w=20$). `Ama.' for Amazon and `Sym.' for Symbotic.}}
\centering
\small
\resizebox{0.9\textwidth}{!}{%
\begin{threeparttable}
\begin{tabular}{c@{\hspace{0.25cm}}l@{\hspace{0.1cm}}|c@{\hspace{0.15cm}}c|c@{\hspace{0.15cm}}c|c@{\hspace{0.15cm}}c}
\toprule\midrule
\multicolumn{2}{c|}{\multirow{2}{*}{{\textbf{Method}}}} & \multicolumn{2}{c|}{\textbf{$N\!=\!80$}} & \multicolumn{2}{c|}{\textbf{$N\!=\!100$}} & \multicolumn{2}{c}{\textbf{$N\!=\!120$}} \\
\multicolumn{2}{c|}{} & \textbf{TPA} $\uparrow$ & \textbf{Time(s)} $\downarrow$  & \textbf{TPA} $\uparrow$ & \textbf{Time(s)} $\downarrow$  & \textbf{TPA} $\uparrow$  & \textbf{Time(s)} $\downarrow$  \\
\midrule
\multirow{2}*{\rotatebox{90}{\textbf{Ama.}}} 
& DQ-RH-PP   & 27.23 $\pm$ 0.77 & 0.17 & 20.89 $\pm$ 1.64 & 0.25 & 17.66 $\pm$ 0.56 & 0.48 \\

& \textbf{RL-RH-PP} ($K=5$)& \textbf{31.80} $\pm$ 0.46 & 0.65 & \textbf{ 28.84} $\pm$ 0.35 & 0.79 & \textbf{25.76}  $\pm$0.55 & 0.96 \\
\midrule
\multirow{2}{*}{\rotatebox{90}{\textbf{Sym.}}} 
& DQ-RH-PP  & 17.33 $\pm$ 0.56 & 0.15 & 14.01 $\pm$ 2.31 & 0.22 & 9.88 $\pm$ 0.74 & 0.43  \\

& \textbf{RL-RH-PP} ($K=5$)  & \textbf{18.38} $\pm$ 0.50  & 0.68 & \textbf{15.38} $\pm$ 1.85 & 0.82 & \textbf{11.31} $\pm$ 2.21& 0.99 \\
\midrule
\bottomrule
\end{tabular}
\end{threeparttable}}
\label{tab:heuristic_compare}
\end{table}

Across all settings, we observe that RL-RH-PP consistently outperforms DQ-RH-PP in terms of throughput per agent. This highlights the effectiveness of our proposed Top-$K$ sampling, demonstrating that the constrained priority order space contains high-quality solutions that can be efficiently retrieved through multiple trials of sampling. These findings suggest that relying solely on a fixed-rule heuristic to construct priority orders may not be an optimal approach for achieving efficient coordination.

\subsubsection{Replace RL with contextual bandit formulation.}

\begin{figure*}[t]
\centering
\includegraphics[width=0.55\textwidth]{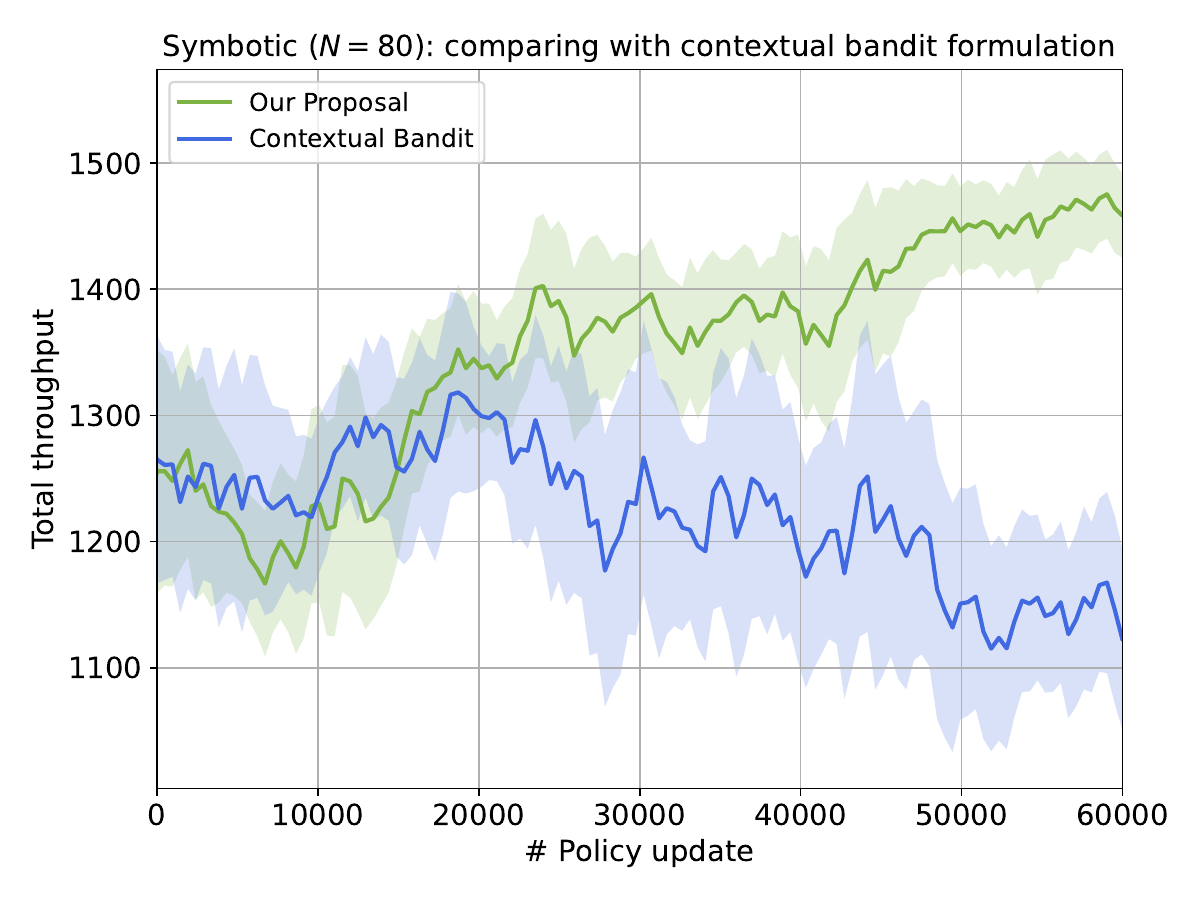}
\caption{\RC{Total throughput during training for contextual bandit ablation study, evaluated on-the-fly with $N=80$}}
\label{fig:ablation_bandit}
\end{figure*}

\RC{To further demonstrate the contribution of multi-step planning, we conduct an ablation study in which the RL agent is trained with a horizon of one step - essentially a contextual bandit formulation. This simplified setup removes the influence of long-term future planning, allowing us to evaluate the performance of the encoder/decoder architecture on its own. As shown in Figure~\ref{fig:ablation_bandit}, the contextual bandit variant converges more quickly, however, its ultimate throughput is lower compared to our RL-RH-PP. This suggests that while the encoder/decoder effectively captures context, the ability to plan over long-horizon using RL is crucial for managing congestion and achieving higher overall performance.}

\section{What Does RL Learn?}
\label{sec:interpretatoin}

\RALL{In this section, we take a closer look at specific planning steps and visually analyze the step-wise priority order sampling of RL-RH-PP to interpret what RL has learned.} 
\hzz{To this end, we analyze both what the policy learns and how it leverages that knowledge to improve planning performance. First, we investigate the priority assignment patterns using heatmaps over time, revealing how RL-RH-PP adapts to congestion through strategic prioritization. Then, we examine the policy’s ability to recover from congested states, comparing movement directions towards next goals derived from shortest paths with those chosen by the RL policy.}

\subsection{\RC{Priority} Heat Maps}
\label{subsubsec:heatmaps}
\begin{figure*}[t]
\centering
\adjustbox{max width=0.85\textwidth, center}{%
    \includegraphics{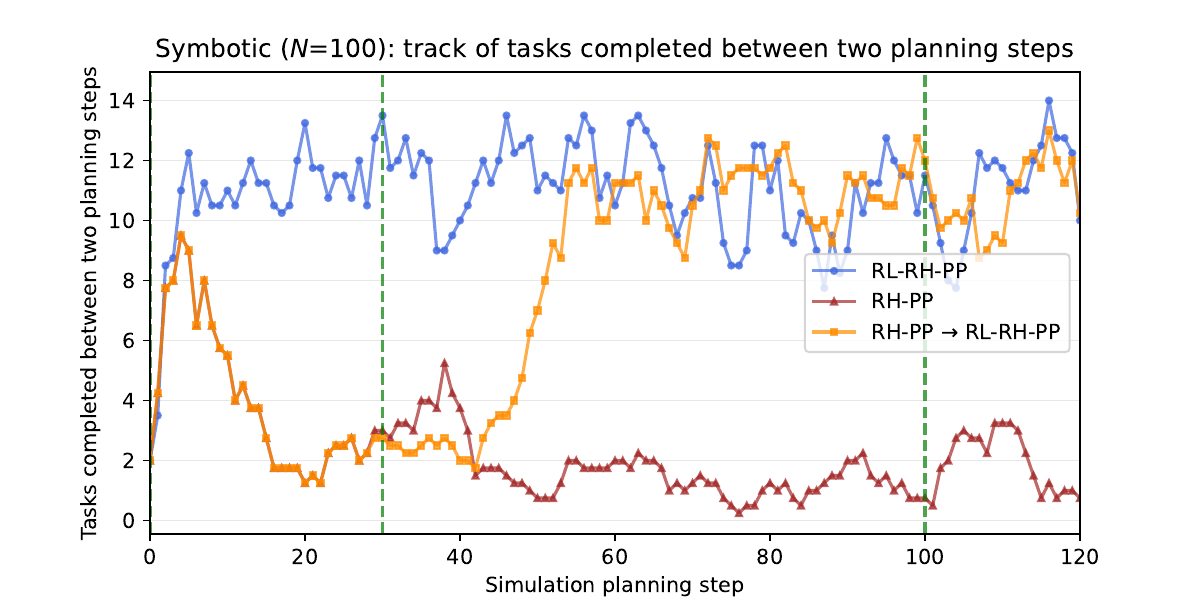}
}
\caption{\RALL{Number of tasks completed during simulation (Symbotic map, $N=100$). We measure how many tasks are completed between consecutive planning steps. The green dashed vertical lines highlight the key planning steps 0, 30, and 100 shown in Figure~\ref{fig:Heatmap}. In the RH-PP $\rightarrow$ RL-RH-PP curve, we first use RH-PP for planning and switch to RL-RH-PP at planning step 30.}}
\label{fig:task_trend}
\end{figure*}

\begin{figure*}[h]
\centering
\adjustbox{center}{%
    \includegraphics[scale=0.9]{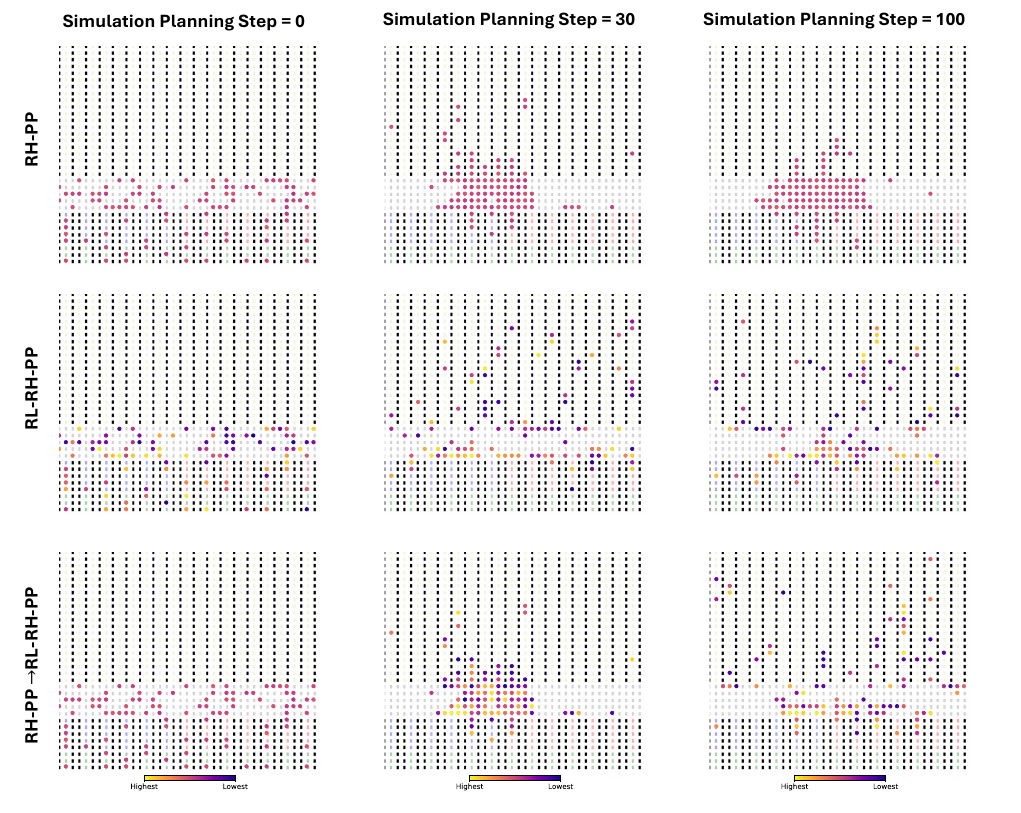}
}
\caption{\RALL{Priority heatmaps for three planning methods (rows) at different steps (columns): 0, 30, and 100, as shown in Figure~\ref{fig:task_trend}. Agents are shown in dots. The top row shows RH-PP, which assigns uniformly random priorities and experiences severe congestion by step 100. The middle row shows RL-RH-PP, which adaptively assigns higher priorities for agents in congested regions, sustaining high throughput. The bottom row shows switching from RH-PP (planning steps 0--29) to RL-RH-PP at step 30, effectively resolving inherited congestion by step 100. Lighter colors indicate higher priority assignments.}}
\label{fig:Heatmap}
\end{figure*}

\hzz{We conduct a step-wise analysis of the tasks completed every $h$ execution steps during simulation in a single evaluation environment for an RL-RH-PP policy trained on Symbotic map with $N=100$ and $w=20$, as illustrated in Figure~\ref{fig:task_trend}. Our observations show that RL-RH-PP consistently maintains a high task-completion rate throughout the simulation. In contrast, RH-PP initially achieves comparable performance but suffers a rapid decline as the simulation progresses and congestion accumulates. This decline suggests that suboptimal priority assignments lead to premature commitments, thereby generating long-term conflicts or deadlocks that hinder effective solutions in subsequent planning steps. Note that the planning step is defined as where the replanning happens. In our setting with $T=800$ and $h=5$, there are a total of 160 planning steps.}

\hzz{To gain further insight into what the RL policy has learned, we pause the simulation at planning steps 0, 30, and 100 (vertical dash lines in Figure~\ref{fig:task_trend}), sampled 100 priority orders from both RH-PP and RL-RH-PP, and compute the average priority assignment for each agent across these samples. As shown in Figure~\ref{fig:Heatmap}, the lighter shades represent higher priority assignments.}

\hzz{As expected, priority assignments from RH-PP (first row in Figure~\ref{fig:Heatmap}) are purely random, resulting in mid-level average priorities across agents. This randomness aligns with the fact that RH-PP samples priorities uniformly, inherently having higher entropy. Additionally, at planning step 100, RH-PP demonstrates noticeable congestion.}

\hzz{Interestingly, as depicted in the second row of Figure~\ref{fig:Heatmap}, RL-RH-PP has an order sampling distribution with lower entropy, and  RL-RH-PP strategically assigns higher priorities to agents located in congested regions as the simulation progresses (steps 30 and 100).  This adaptive prioritization implies that the RL policy has implicitly learned to recognize and proactively manage potential bottlenecks. By assigning higher priorities to agents within congested areas, RL-RH-PP effectively mitigates conflict propagation, reducing the likelihood of deadlocks and improving overall system throughput. This learned behavior is crucial, as it directly addresses congestion at its onset, providing long-term stability and more efficient resource utilization during subsequent planning stages.}

\hzz{From Figure~\ref{fig:task_trend} and the first row of Figure~\ref{fig:Heatmap}, we observe that RH-PP begins to accumulate congestion around planning step 30. To investigate whether the RL policy can effectively recover from such congested states, we conduct an experiment where RH-PP is used for the first 29 planning steps, followed by switching to RL-RH-PP for the remaining steps starting from planning step 30. This setup corresponds to the RH-PP $\rightarrow$ RL-RH-PP curve in Figure~\ref{fig:task_trend} and the third row of Figure~\ref{fig:Heatmap}. Interestingly, we find that task completion per planning step starts to improve shortly after the switch, indicating that the congestion is being alleviated. Moreover, the heatmap at planning step 100 in the third row shows a significantly reduced level of congestion compared to RH-PP alone (first row), further confirming this recovery. These results suggest that the learned RL policy is capable of proactively managing congestion and effectively resolving severe deadlocks when inherited from suboptimal priority planning,} \hzzz{despite not being explicitly trained to do so.}

\begin{figure*}[h]
\centering
\adjustbox{center}{%
    \includegraphics[scale=0.75]{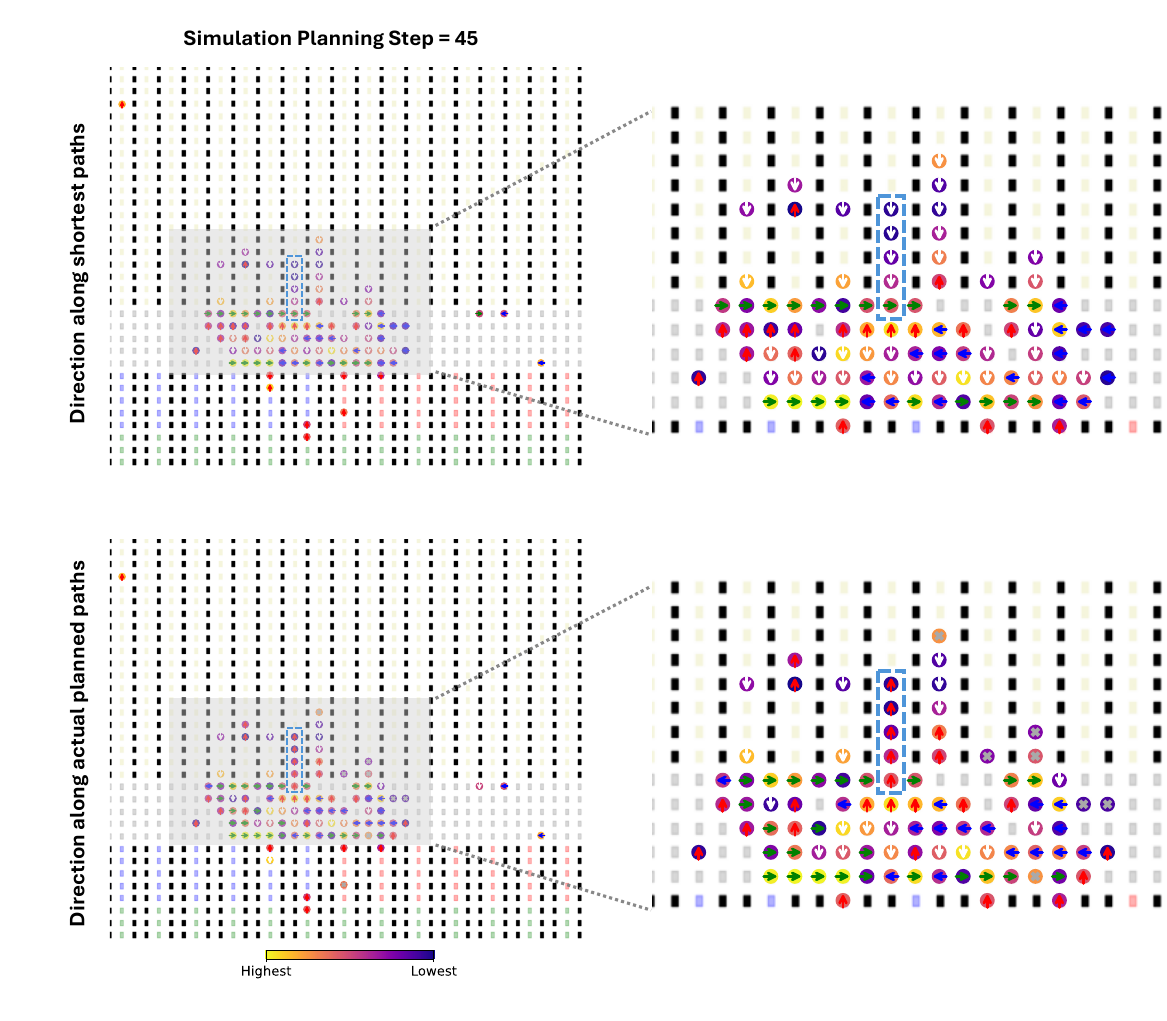}
}

\caption{\RALL{Next movement directions along shortest path and the actual planned path of RH-PP $\rightarrow$ RL-RH-PP curve at planning step = 45 in Figure~\ref{fig:task_trend}. The blue bounding boxes highlight the agents located near the boundary of the congested region. Different ordinal directions are shown in different colors, the cross symbol indicates \textit{wait} action.}}
\label{fig:step direction}
\end{figure*}

\subsection{\hzz{How Does RL-RH-PP Recover from Congestion}}

\hzz{The evidence of RL-RH-PP successfully recovering from the congestion built up by RH-PP, as discussed in Section~\ref{subsubsec:heatmaps}, is fascinating. To gain a deeper understanding of how the RL policy facilitates this recovery, we analyze the behavior of RL-RH-PP at planning step 45-the point at which the RH-PP $\rightarrow$ RL-RH-PP curve in Figure~\ref{fig:task_trend} begins to rise, indicating that congestion is starting to resolve. Specifically, we compare the hypothetical next movement direction toward each agent's next goal, derived from their shortest paths, with the actual next movement direction selected by RL-RH-PP. These directions are visualized as arrows in Figure~\ref{fig:step direction}.}

\hzzz{Interestingly, we observe a key behavioral shift in the RL-guided planning. For agents located near the boundary of the congested region (highlighted in the blue bounding boxes), we see that the actual planned direction given by RL-RH-PP tends to be counterintuitive - away from the hypothetical shortest-path directions which guide agents to the goals. This “backtracking” behavior emerges most clearly when agents deeper within the congestion must pass through a narrow aisle that is currently blocked (see the vertical blue bounding box). By recognizing that such agents become space-time obstacles for those agents closer to the aisle entrance, RL-RH-PP proactively repositions these boundary agents by assigning them lower priorities to clear the route. Specifically, RL-RH-PP may temporarily reverse their direction, opening a passage so that the congested agent can advance without colliding or creating a bottleneck. Once the way is cleared, these boundary agents resume their forward progress toward their original goals. Such behavior demonstrates that the RL policy has learned to coordinate local detours that reduce global conflict, enabling more efficient resolution of congestion and improving overall throughput.}

\begin{figure*}[t]
\centering
\adjustbox{center}{%
    \includegraphics[scale=0.35]{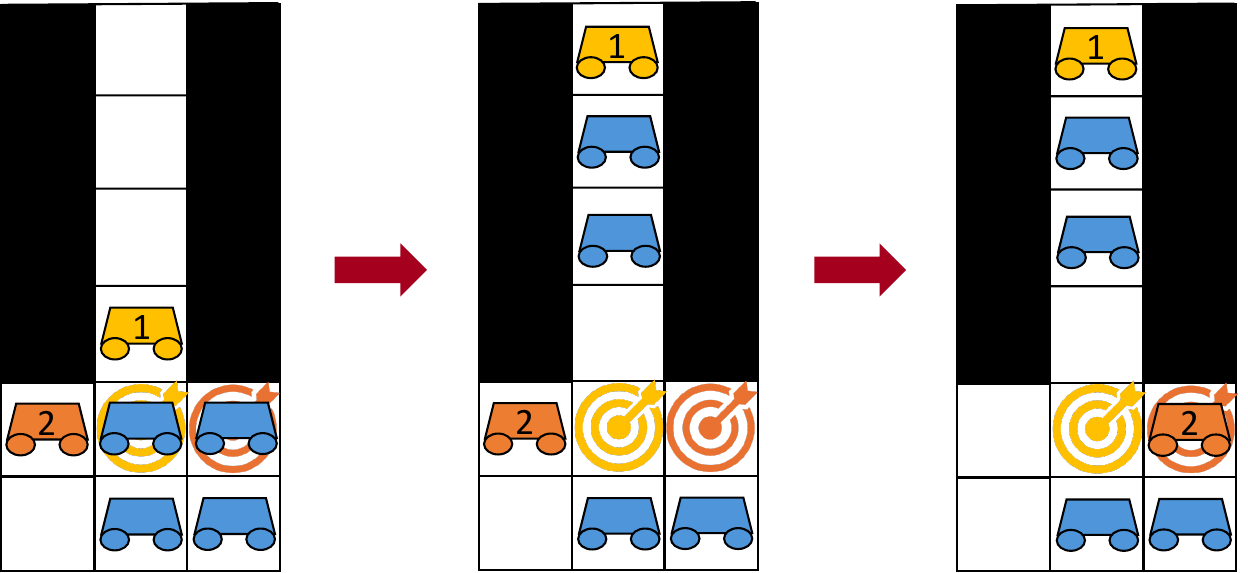}
}

\caption{\hzzz{Illustrative grid world example of RL-RH-PP congestion recovery. It shows a deck and a narrow aisle with agents labeled “1” (lower priority) and “2” (higher priority), and their goals. As a result, agent 1 backtracks to clear the path, enabling agent 2 to move forward. This temporary detour mitigates congestion and prevents deadlocks, thereby improving throughput.}}
\label{fig:example_grid}
\end{figure*}

\hzzz{To further clarify the mechanism behind RL-RH-PP’s recovery from congestion, consider a hypothetical grid world example in Figure~\ref{fig:example_grid}. In this scenario, agents are positioned in a congested desk with a narrow aisle. We specifically focus on agents 1 and 2. Initially, both agents appear to be heading toward their respective goals. The RL-RH-PP policy assigns a higher priority to agent 2, which is located deeper within the congested area, and a lower priority to agent 1, positioned near the aisle’s exit. Under such a priority assignment, although agent 1 is closer to its destination, its action is not to move forward but to momentarily backtrack. By stepping backward, agent 1 clears the path for other congested agents (blue color) and lets the high-priority agent 2 pass, effectively turning itself from a potential blockage into an enabler of smoother flow. Once agent 2 passes the aisle's exit, agent 1 can resume its forward progress, and the deadlock is resolved.}

\section{Conclusion}
\label{sec:conclusion}
In this paper, we introduced RL-guided Rolling Horizon Prioritized Planning (RL-RH-PP), the first framework that integrates reinforcement learning with a classical search-based planner for learning-guided optimization of lifelong MAPF. Our approach builds upon a rolling-horizon extension of Prioritized Planning (RH-PP) to continuously replan in high-density warehouse environments. By casting dynamic priority ordering as a Partially Observable Markov Decision Process, we train a transformer-based neural network that autoregressively decodes high-quality priority orders, effectively reducing the search space for the single-agent path solver.

\RALL{Through extensive experiments in two large-scale warehouse environments, inspired by Amazon and Symbotic, we observe that integrating RL into the RH-PP backbone delivers an average 25\% improvement across both simulators, which in turn enables RL-RH-PP to achieve higher throughput than conventional lifelong MAPF baselines. Notably, RL-RH-PP consistently outperforms RH-PP with random sampling, and showcases unique benefits over strong baselines such as RH-CBS, RH-PBS, PIBT, and WPPL, especially in highly constrained scenarios. Moreover, our learned policy exhibits strong zero-shot generalization across varying agent densities, planning horizons and warehouse layouts. \RB{This adaptability is facilitated by a learnable dictionary-based position embedding and attention-based neural architecture that captures both spatial and temporal inter-agent dependencies.}
We additionally report thorough ablation studies verifying the importance of our core design choices. Finally, a case study demonstrates that the RL component learns effective congestion-avoidance policies and can adaptively recover feasible plans when congestion has already arisen under RH-PP. This evidences a unique and potentially impactful practical value of our proposed learning-guided optimization approach in handling long-term dynamics, particularly in complex systems such as warehouse automation.
}

\RALL{
Our work also highlights the promise of learning-based approaches that complement, rather than replace, established search-based solvers. Our RL-RH-PP effectively capitalizes on the strengths of both data-driven decision-making and efficient single-agent path searches. Moreover, some experiments suggest that RL-RH-PP can reverse the performance decline of RH-PP under challenging conditions, indicating that our approach may help address inherent limitations of the backbone solver to some extent, which is likely due to the ability of reinforcement learning to automate data-driven coordination and prioritization. These findings underscore the potential for learning-guided hybrid methods to achieve state-of-the-art performance in the complex, dynamic, and fast-evolving real-world applications such as multi-robot coordination in warehouse automation.}

\RB{One promising avenue for future research is to scale our approach to even larger warehouses with thousands of agents by parallelizing the evaluation of multiple candidate priority orders. While Top-K sampling can yield high-quality solutions, it may increase computational overhead. Efficient, parallelizable planning engines could mitigate this cost, making RL-RH-PP feasible for real-time operation with thousands of robots.} \RB{Meanwhile, our encoder indexes absolute map locations. Such a representation design allows zero-shot transfer to layout variation with fixed map size, but precludes zero-shot transfer to layouts with different map sizes. To broaden applicability, future work will pursue full map-agnostic generalization state representations, training regimes that encourage cross-map-size generalization.} \RC{We also note that our neural architecture for generating priority orders could, in principle, be applied to the one-shot MAPF setting; one might design an alternative learning paradigm to train it for producing effective static orders, though we leave this as a separate line of research beyond the scope of this paper.} Moreover, our framework naturally extends to more complex warehouse tasks. For example, jointly optimizing task assignment and path planning could yield a fully integrated solution for maximizing throughput.  \RA{Concretely, the autoregressive decoder can emit a sequence of \((\text{agent},\text{task})\) pairs rather than an agent-only priority order. At each decoding step, we apply masks to enforce feasibility (each task assigned at most once, agent queue/capacity limits, task release/eligibility), append the selected task to the agent's queue, and continue until a stopping criterion is met (e.g., budgeted number of pairs or no feasible pairs). The resulting partial assignment for the window is then executed by the same RH-PP planner. Training can optimize a combined throughput/tardiness reward.} Incorporating robustness to model uncertainties, such as stochastic delays or partial observability, would further enhance real-world applicability. \RALL{Finally, we believe it is promising to explore similar learning-guided optimization frameworks for addressing long-horizon dynamics in broader classes of optimization and decision-making problems.}


\begin{acks}
  This research was supported by Symbotic and a Fundamental Research grant from the National Science Foundation and AI Singapore (NSF-AISG).
\end{acks}

\bibliographystyle{ACM-Reference-Format}
\bibliography{sample}

\newpage
\appendix

\section{Mixed Zero-Shot Transfer to Different Agent Numbers, Planning Horizon, and Map Layouts.}
\label{apd: mixed_agent_horizon}

\begin{figure*}[h]
\centering
\includegraphics[width=0.8\textwidth]{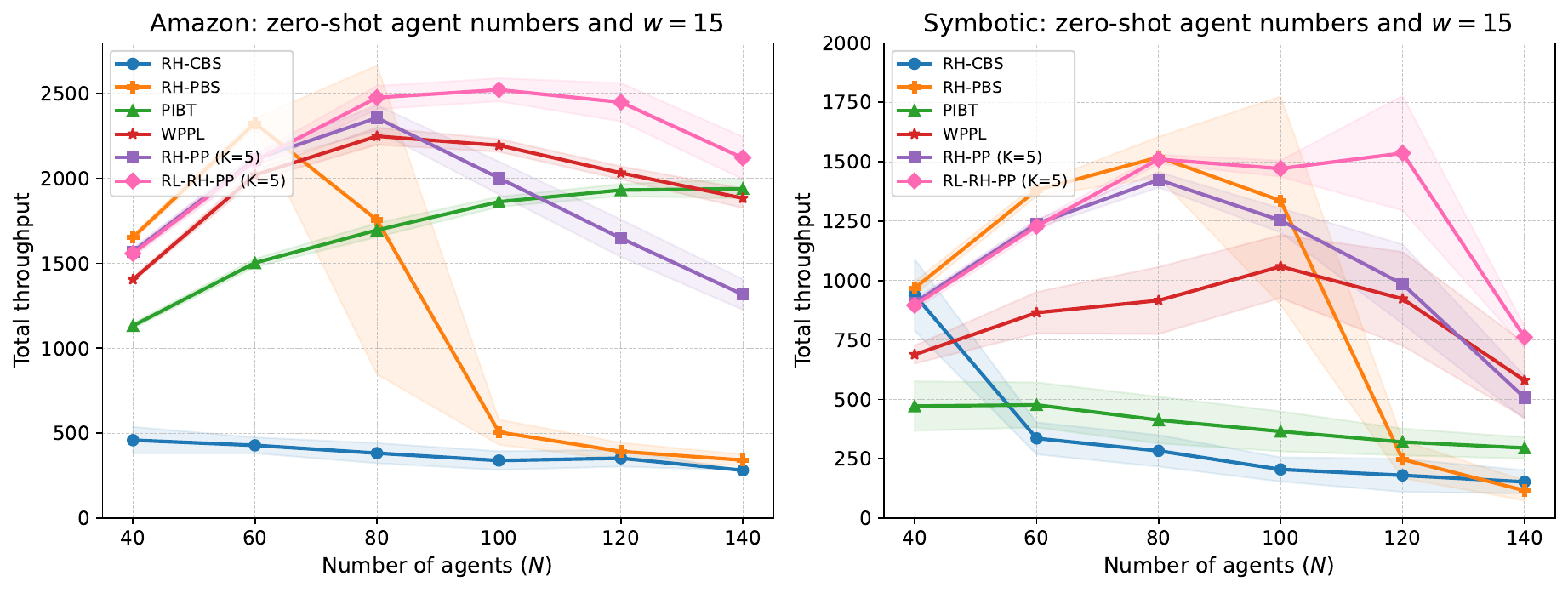}
\caption{\RABC{Total throughput versus agent number $N$ in zero-shot transfer evaluation. The base model is trained at $N=120$ and $w=20$, and evaluated zero-shot at other $N$ with $w=15$.}}
\label{fig:zeroshot15}
\end{figure*}

\begin{figure*}[h]
\centering
\includegraphics[width=0.8\textwidth]{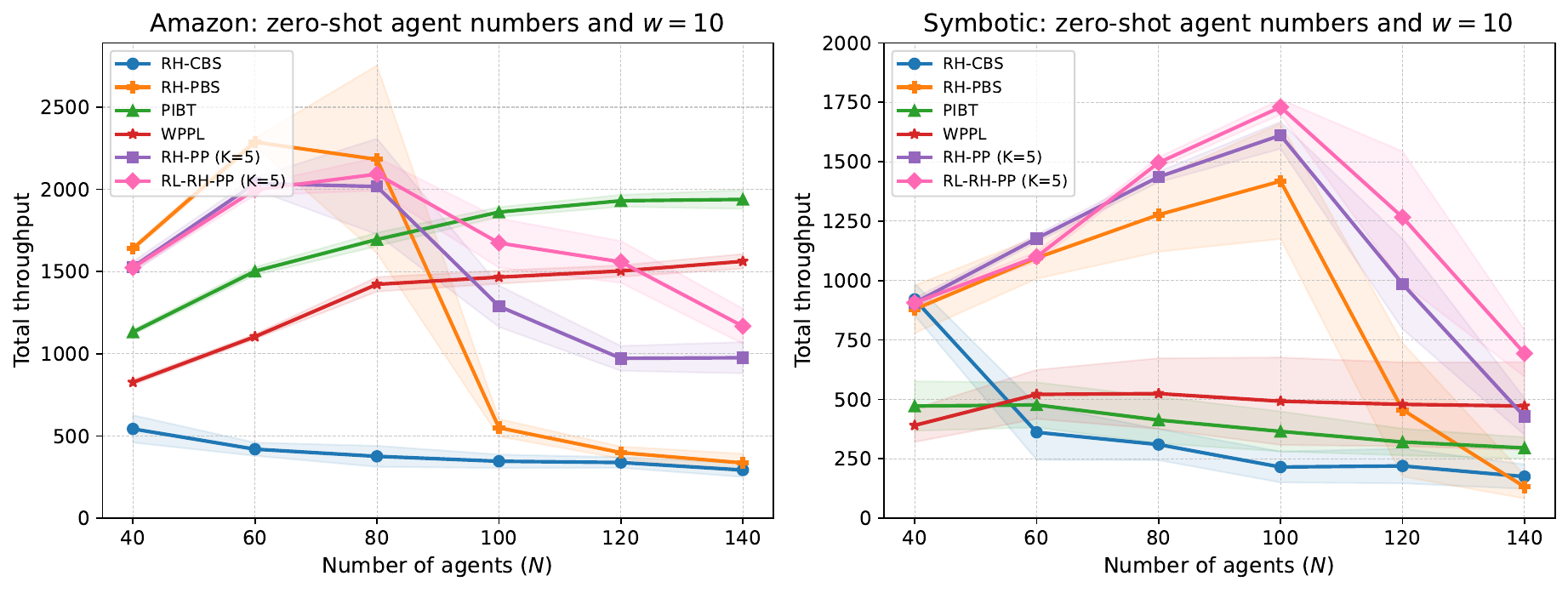}
\caption{\RABC{Total throughput versus agent number $N$ in zero-shot transfer evaluation. The base model is trained at $N=120$ and $w=20$, and evaluated zero-shot at other $N$ with $w=10$.}}
\label{fig:zerpshot10}
\end{figure*}


\begin{figure*}[h]
\centering
\includegraphics[width=\textwidth]{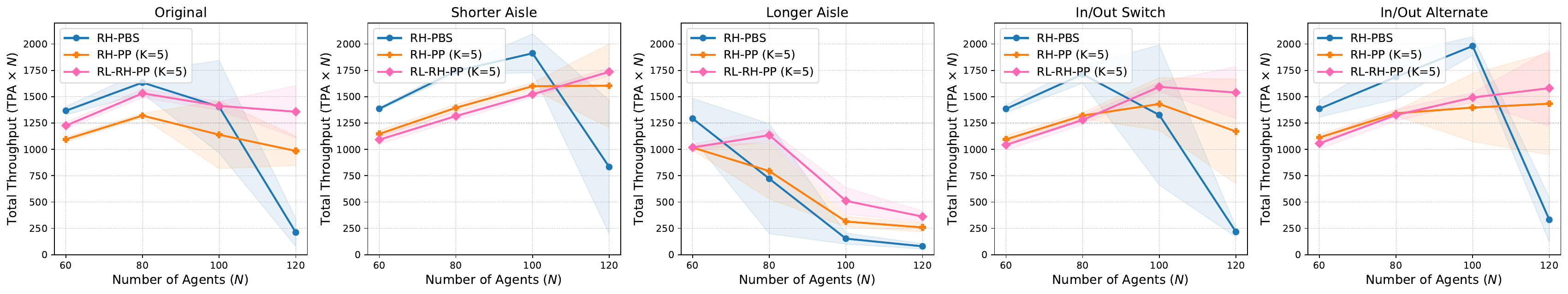}
\caption{Total throughput versus different Symbotic maps in zero-shot transfer evaluation. The base model is trained at $N=120$ and $w=20$ on the Original map, and evaluated zero-shot at other map layouts with $w=20$.}
\label{fig:mixed_agent_layout_apd}
\end{figure*}

\section{Supplementary Results for $\beta$ and $K$ in the Ablation Study.}
\label{apd: beta}

\RABC{Similar effects of increasing $\beta$ and $K$ is observed compared to the result shown in our main manuscript.}

\begin{figure*}[h]
\centering
\includegraphics[width=0.85\textwidth]{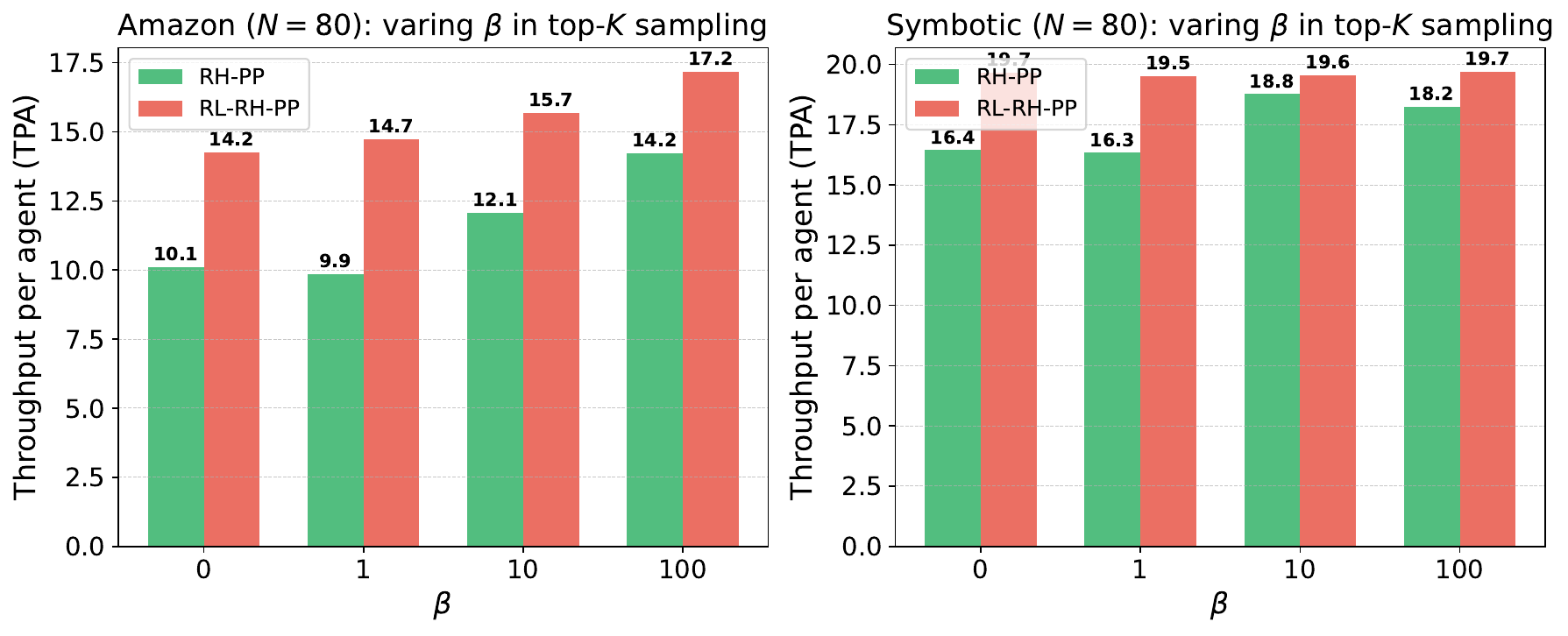}
\caption{\RB{Throughput per agent (TPA) at evaluation vs $\beta$, evaluated with $N=80$}}
\label{fig:TBA_beta_amazon_symbotic_80_apd}
\end{figure*}

\begin{figure*}[t]
\centering
\includegraphics[width=0.85\textwidth]{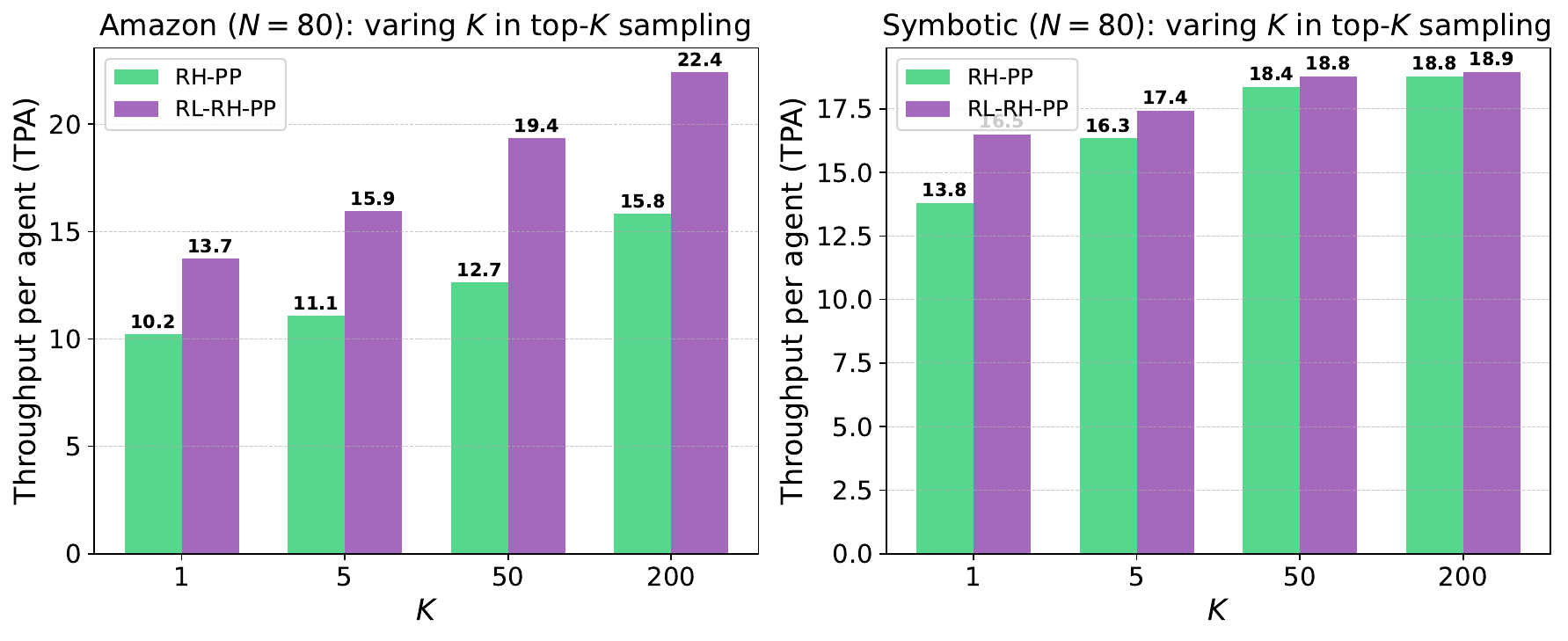}
\caption{\RABC{Throughput per agent (TPA) at evaluation vs $K$, evaluated with $N=80$}}
\label{fig:TBA_K_amazon_symbotic_80_apd}
\end{figure*}

\section{\hzz{Notation Table}}
\vspace{30pt}

\begin{longtable}{l p{0.65\textwidth}} 
\toprule 
\textbf{Notation} & \textbf{Description} \\ 
\midrule 
\endfirsthead

\multicolumn{2}{c}%
{\tablename\ \thetable\ (continued)} \\[6pt] 
\toprule 
\textbf{Notation} & \textbf{Description} \\ 
\midrule 
\endhead

\endfoot

\bottomrule 
\endlastfoot

\multicolumn{2}{l}{\textbf{General MAPF and Agent Parameters}} \\ \midrule
$N$ & Number of agents. \\
$V, E$ & Vertices and edges defining the MAPF graph $G=(V,E)$. \\
$W, H$ & Width and height of the warehouse map. \\
$g_i$ & Goal location of agent $i$. \\
$o_i$ & Indicator if agent $i$ is forced to use its shortest path (when no feasible single-agent path is found). \\
$\pi_i$ & Path sequence of locations for agent $i$. \\
$d_{i,t}$ & Average distance of agent $i$ to its next goals at time $t$. \\
$\text{TPA}$ & Throughput per agent over the simulation horizon. \\
$t$ & Discrete simulation time step index. \\
$T$ & Total simulation horizon (e.g., 800 simulation time steps). \\
$c_{i,t}$ & binary indicator equal to 1 if, at time $t$, agent $i$'s plan consists of \textit{wait} actions for the next $h$ steps (and 0 otherwise). \\
$s_{i,t}$ & binary indicator equal to 1 if, at time $t$, RH-PP fails to find a feasible path for agent $i$ under the selected priority order (and 0 otherwise). \\

\midrule
\multicolumn{2}{l}{\textbf{Rolling-Horizon Planning}} \\ \midrule
$h$ & Execution horizon in rolling-horizon planning. \\
$w$ & Planning horizon in rolling-horizon planning. \\
$\prec$ & A total priority order over agents. \\
$K$ & Number of candidate priority orders in Top-$K$ sampling. \\
$\mathrm{cost}(\prec_k)$ & Heuristic cost for the $k$-th sampled priority order. \\
$\beta$ & Penalty weight in the cost function $\mathrm{cost}(\prec_k)$. \\
$\kappa$ & Weighting factor for congestion penalty in the reward. \\
$\sigma$ & Weighting factor for infeasibility penalty in the reward.\\

\midrule
\multicolumn{2}{l}{\textbf{Reinforcement Learning and PPO}} \\ \midrule
$a_t$ & Action at simulation time step $t$, representing the selection of one or more priority orders. \\
$\boldsymbol{o}_t$ & Observation at time $t$. \\
$R(\boldsymbol{o}_t, a_t)$ & Reward at observation $\boldsymbol{o}_t$ and action $a_t$. \\
$\sigma_{i}^t$ & shortest path of agent $i$ at simulation time step $t$. \\
$r$ & (maximum) length of the shortest path.\\
$A(\boldsymbol{o}_t, a_t)$ & Advantage function in PPO. \\
$\hat{R}_t$ & Reward-to-go from simulation time step $t$ onward in PPO. \\
$\gamma$ & Discount factor in reinforcement learning. \\
$\epsilon$ & PPO clipping parameter. \\
$\eta$ & Entropy coefficient in PPO’s objective. \\
$f$ & Rollout reuse per epoch in PPO (reusing collected data for multiple updates). \\
$B$ & Batch size for PPO updates. \\

$Z$ & Total number of PPO training epochs (policy updates). \\
$\tau$ & Rollout trajectories in PPO training. \\
$\mathcal{D}$ & Rollout dataset. \\
$\alpha$ & Adam learning rate (used interchangeably with $\lambda$). \\
$\omega$ & Learning rate decay factor (exponential). \\
$g_{\mathrm{clip}}$ & Gradient clipping threshold. \\
$\phi, \theta$ & Parameters of the value function and policy 
networks in RL. \\

\midrule
\multicolumn{2}{l}{\textbf{Neural Architecture}} \\ \midrule
$L$ & Number of stacked encoder layers in the transformer. \\
$U$ & Number of attention heads in each multi-head attention layer. \\
$u$ & The $u$-th head in attention. \\
$d$ & Embedding dimension used in the network (e.g., for position embeddings). \\
$\mathbf{H}^{(\ell)}$ & Hidden representation output by the $\ell$-th encoder layer. \\
$\mathbf{X}$ & Dictionary of learnable position embeddings for each cell/vertex. \\
$\mathbf{K}, \mathbf{V}$ & Key and value matrices used in the transformer model. \\
$\mathbf{L}$ & Logit key matrix used for sampling in the decoder. \\
$\mathbf{W}^{K}, \mathbf{W}^{V}, \mathbf{W}^{L}$ & Learnable projection matrices for keys, values, and logit keys. \\
$\mathbf{W}^{Q_{\text{step}}}$ & Learnable projection matrix for query at decoding step. \\
$\mathbf{f}$ & Fixed global context embedding. \\
$p$ & Autoregressive decoding step index in the transformer decoder. \\
$a_n$ & Selected agent at decoding step $n$. \\
$\mathcal{U}_{n-1}$ & Set of agents already chosen in previous decoding steps. \\
$p(a_n = i \mid \mathbf{q}_n, \mathcal{U}_{n-1})$ & Probability of selecting agent $i$ at decoding step $n$. \\
$\mathbf{q}_n$ & Query vector at decoding step $n$. \\
$\mathbf{q}_{n,u}$ & Query vector at decoding step $n$ for attention head $u$. \\
$\mathbf{g}_n$ & Glimpse vector at decoding step $n$. \\
$\boldsymbol{\rho}_n$ & Unnormalized logits at decoding step $n$. \\

\end{longtable}

\section{\hzz{Experiment Details}}

\begin{table}[t]
\centering
\caption{Hyperparameter Settings for RL-RH-PP}
\label{tab:hyperparam_table}
\begin{tabular}{l l}
\toprule
\textbf{Parameter} & \textbf{Value} \\
\midrule
Training epochs             & 4000  \\
Rollout reuse per epoch ($f$)       & 3     \\
Entropy loss weight ($\eta$)        & 0.01  \\
PPO clipping parameter ($\epsilon$) & 0.2   \\
Initial learning rate ($\lambda$)   & 0.001 \\
Learning rate decay ($\omega$)      & 0.999 \\
Batch size ($B$)                    & 32    \\
Gradient clipping threshold ($g_{\mathrm{clip}}$) & 0.5 \\
Discount factor ($\gamma$)          & 0.99  \\
Number of encoder layers ($L$)      & 2     \\
Number of attention heads ($H$)     & 4     \\
Embedding dimension ($d$)           & 32   \\
\bottomrule
\end{tabular}
\end{table}

For the selection of hyperparameters in RL training, please refer to Table~\ref{tab:hyperparam_table}. \hzz{Each parameter is chosen based on 5 empirical trials (considering memory usage, rollout efficiency, and training stability) rather than an extensive hyperparameter search or optimization. Additionally, we report an average environment sampling rate of 0.064 CPU time per parallel simulation step, achieved using an AMD Ryzen Threadripper PRO 5995WX CPU with 128 cores, running multi-threaded parallel simulations across 16 environments.}


\FloatBarrier
\vspace*{2\baselineskip}
\end{document}